\documentclass{article}

\usepackage[preprint]{nips_2018}

\usepackage{xcolor}

\usepackage[utf8]{inputenc} 
\usepackage[T1]{fontenc}    
\usepackage{hyperref}       
\usepackage{url}            
\usepackage{booktabs}       
\usepackage{amsfonts}       
\usepackage{nicefrac}       
\usepackage{microtype}      
\usepackage{graphicx}
\usepackage{subcaption}
\usepackage{amssymb}
\usepackage{amsmath}
\usepackage{bm}

\newcommand{\dcgan}{\texttt{1}}
\newcommand{\wgan}{\texttt{3}}
\newcommand{\feature}{\texttt{2}}
\newcommand{\conditional}{\texttt{4-cond}}
\newcommand{\eleven}{\texttt{5-cond}}
\newcommand{\pixelcnn}{\texttt{6-auto}}

\title{Skill Rating for Generative Models}

\author{
  Catherine Olsson, Surya Bhupatiraju, Tom Brown, Augustus Odena, Ian Goodfellow\\
  Google Brain\\
  \texttt{\{catherio, sbhupatiraju, tomfeelslucky, augustusodena, goodfellow\}@google.com}
}

\begin{document}

\maketitle

\begin{abstract}
We explore a new way to evaluate generative models using
insights from evaluation of competitive games between human players.
We show experimentally that tournaments between generators and discriminators
provide an effective way to evaluate generative models.
We introduce two methods for summarizing tournament outcomes:
tournament win rate and skill rating.
Evaluations are useful in different contexts, including monitoring the
progress of a single model as it learns during the training process,
and comparing the capabilities of two different fully trained models.
We show that a tournament consisting of a single model playing against
past and future versions of itself produces a useful measure of
training progress.
A tournament containing multiple separate models (using different seeds,
	hyperparameters, and architectures)
provides a useful relative comparison between different
trained GANs.
Tournament-based rating methods are conceptually distinct from numerous
previous categories of approaches to evaluation of generative models,
and have complementary advantages and disadvantages.
\end{abstract}
\section{Introduction}

Evaluation of generative models is a difficult task.
Many conceptually different approaches have been explored, each with significant disadvantages.
See \citet{Theis2015d} and \citet{Borji2018} for an overview of these approaches and demonstrations of their shortcomings.

We propose a new framework for evaluating generative models via an adversarial
process, in which many models compete in a tournament.
We leverage evaluation methodologies developed previously for the evaluation
of human competitors to quantify performance in such tournaments.

In games such as chess or tennis, skill rating systems such as Elo \citep{Elo1978} or Glicko2
\citep{Glickman2013} evaluate players by observing a record of wins and losses of multiple
players and inferring the value of a latent, unobserved skill variable for each player
that explains the records of wins and losses. Similarly, we frame the evaluation of generative models as a latent skill estimation problem
by constructing multiplayer tournaments that generalize the two-player distinguishability
game used by noise contrastive estimation (NCE) and generative adversarial networks (GANs)
\citep{Gutmann+Hyvarinen-2010,Goodfellow-et-al-NIPS2014-small,Goodfellow-ICLR2015} and
estimating the latent skill of generative models that participate in these
tournaments.
Each player in a tournament is either a {\em discriminator} that attempts to distinguish
between real and fake data or a {\em generator} that attempts to fool the discriminators
into accepting fake data as real.
While the framework was designed primarily with GANs in mind, we can estimate the skill of
any model capable of playing one of these roles. For example, any model capable of generating samples can
participate as a generator, such as an explicit density model.

We introduce two methods for summarizing tournament outcomes (see Section~\ref{methods}):

\begin{enumerate}
\item Tournament win rate: each generator's average rate of successfully fooling the set of discriminators in the tournament (Section~\ref{winrate})
\item Skill rating, in which a skill rating system (such as the Elo score commonly used for chess rankings, or a related system such as Glicko2) is applied to the tournament outcomes to produce a skill rating for each generator (Section~\ref{skillrating}).
\end{enumerate}

We show experimentally that tournament results provide an effective way to
evaluate generative models.
First, we show that a within-trajectory tournament --- between snapshots of
a single GAN's own discriminator and generator at successive iterations
throughout training --- provides a useful measure of training progress, even
without access to generators or discriminators other than the one being trained
(Section~\ref{monitor}).
Second, we show that a more general tournament --- between generator and
discriminator snapshots from GANs with different seeds, hyperparameters, and
architectures --- provides a useful relative comparison between different
trained GANs (Section~\ref{compare}).

In Section~\ref{context} we place place our work in the larger context of evaluation systems for generative models, and elaborate on the strengths and limitations of our method compared to others.
In Section~\ref{apples} we provide preliminary evidence that
our method is applicable to datasets that are not well-represented by a standardized image embedding, such as unlabeled datasets or modalities other than natural images.
We also show
that using skill rating systems to summarize tournaments makes it possible to skill rate all $n$ players in a tournament without needing to run $n^2$ matches. In Section~\ref{compare} we show that GAN discriminators can successfully judge samples from generators they have not trained against, including other GAN generators and other types of generative models. In Section~\ref{chekhov} we show that our method can be applied even in settings where the generator is nearly perfect.

\section{Context and Related Work}
\label{context}

Accurate evaluation of generative models is necessary to guide research efforts
to improve these models.
However, it is both conceptually difficult to specify what we want from a
generative model in quantitative terms and computationally difficult to
compute the value of many evaluation metrics.
Our tournament-based metrics are computationally tractable and are conceptually
distinct from existing approaches to evaluation, offering a complementary set
of advantages and disadvantages.

One common metric is to report the log-likelihood $\log p_{\rm{model}}(\mathbf{x})$ that a model assigns to test data points $\mathbf{x}$,
or to estimate the likelihood using samples \citep{Breuleux+al-TR-2010}.
These approaches have numerous practical limitations, which have been described by \citet{Theis2015d} and others.
The most common alternative to likelihood is to evaluate some notion of sample
quality.
One
example of this approach is to report assessments by
human raters \citep{denton2015deep,Salimans2016}.
However, this process can yield results that are not reproducible, since
different populations of human raters make different judgments.
For example, \citet{Salimans2016} found that deep learning
researcher Alec Radford had nearly perfect ability to detect GAN samples,
even though the same samples fooled many Mechanical Turk workers.
Additionally, different subpopulations of crowdworkers will accept
different tasks depending on the task structure and pay, and the community
reputation of the requester \citep{Silberman2015}.

Many other approaches to sample quality rating are ad hoc methods of assessing
highly specific problems with generative models. For example, Inception Score
assesses the ability of a model to generate a wide variety of recognizable
classes \citep{Salimans2016} but ignores all other aspects of the generated
samples. Some other metrics focus on the diversity of samples generated by
the model \citep{arora2018do,santurkar2018a}.
Another approach to evaluation is to use a generative model as a component
in an end-to-end system and evaluate performance of the system as a whole.
For example, \cite{Salimans2016} use GAN samples to train semi-supervised
classifiers, and use accuracy in the classification task as a metric.
Finally, another category of metrics is based on moment matching, measuring
the difference in statistics between real data and generated data.
The main example of this in use today is Fr\'echet Inception Distance (FID)
\citep{Heusel2017}.
Moment matching methods must specify which statistics to collect and how
to measure the distance between them; FID uses the mean and covariance of
the last-layer features of an Inception-v3 network \citep{Szegedy-et-al-2015}
and measures the Fr\'echet distance between the Gaussian distributions
defined by these means and covariances.
The main conceptual downside to moment matching methods is that they depend
on the choice of moments; Inception provides a good feature space for images
but it is not clear that similar feature spaces are readily available for
other kinds of data that have not benefited from large labeled datasets
and years of intense study.

To this set of conceptual approaches, we introduce skill rating, based on
the principle of estimating latent skill in games of producing and detecting
fakes. Skill rating systems such as Elo \citep{Elo1978} and TrueSkill \citep{Herbrich2007} have been applied in the evaluation of game-playing systems \citep{Silver2016,Openai2017}, but to our knowledge ours is the first application to generative models.

Our approach complements likelihood because it is computationally tractable
and defined for generator models that offer no density function.
Our approach can compare models that use different input
or output formats (such as continuous versus discrete representations of
image pixels), because reformatting the data does not require an adjustment to the
score.
Our approach is more reproducible than human evaluation and captures more
aspects of the data than ad hoc methods aimed at measuring single properties
such as sample diversity.
Finally, our method is more adaptable than moment matching approaches, because
it does not require the experimenter to specify a fixed feature set; players
in the tournament can learn to attend to any features that are useful to win.

Some downsides to our approach include that it provides a relative rather than
absolute score of a model's ability, that tournaments among many types of models
involve greater software complexity than other metrics, and that reproducing
scores requires reproducing the population of models used in the tournament.
We provide evidence that GAN discriminators can successfully
judge samples from generators other than the one they trained against; however,
this is an empirical claim that our method works in practice, and we do not
attempt a theoretical explanation for when and why out-of-distribution discrimination should be expected to work. (See Section~\ref{chekhov} for an exploration of our method's behavior when the
generator has trained to near-perfect performance).
Finally, the specific format of tournament we used in this paper involves games played
over single samples, so generators that suffer from low diversity can perform
well in our tournaments, but this could be resolved in future work by developing
tournaments that involve games played at the batch level.

The most closely-related metric to our tournament-based approach
are the generative adversarial metric (GAM) \citep{im2016generating},
and the generative multi-adversarial metric (GMAM) \citep{Durugkar2017} which is an extension to
generators whose training process involved multiple discriminator networks.
GAM involves a competition between two generators and two discriminators.
GAM requires both discriminators to have roughly equal performance on a fixed
test set, otherwise the match is declared a tie (eq 11).
GAM declares which of two generators is better, but does not assign a numeric
rating comparable across many different models.
GAM is able to assess non-GAN generators.
Our approach involves tournaments potentially much larger than two models.
We do not require discriminators to have roughly equal skill in any way;
we are able to rank the skill of both generator and discriminator players
by observing the outcome of many different matches.
We assign quantitative scores that allow many different models to be compared
with the same scoring system, not just a determination that one model is
better than one other model.

Overall, we believe that our new conceptual approach of latent skill estimation
holds promise as a complementary evaluation technique.

\section{Methods}
\label{methods}

\begin{figure}
  \centering
  \includegraphics[width=1.0\linewidth]{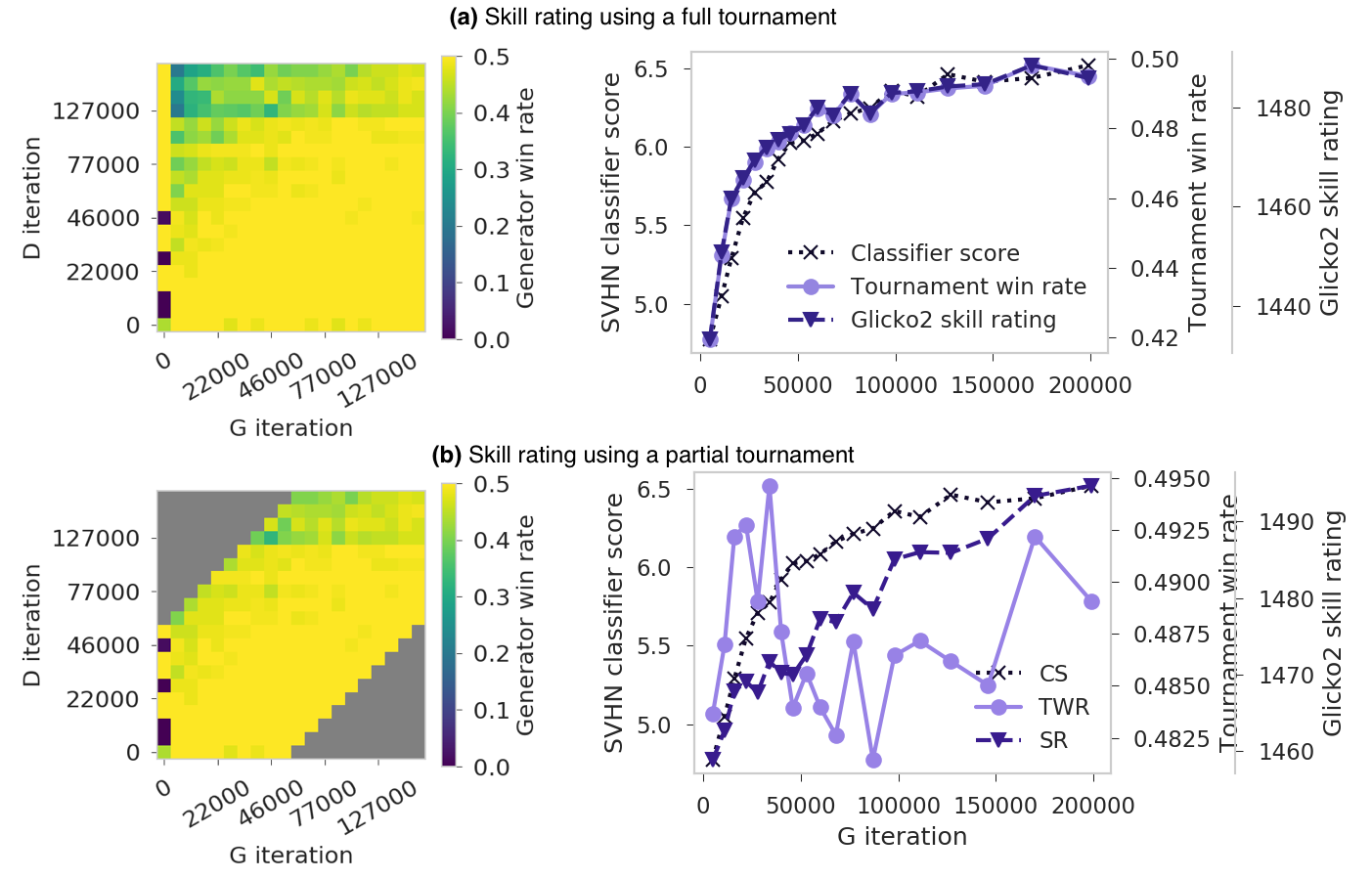}

    \caption{\textbf{Within-trajectory tournament outcomes for experiment \dcgan{}.} In the upper half of the figure: Figure~\ref{fig-monitor}a-left shows raw tournament outcomes. Each pixel represents the average win rate between one generator and one discriminator from different iterations of experiment \dcgan{}. Brighter pixel values represent stronger generator performance. Figure~\ref{fig-monitor}a-right compares tournament summary measures to SVHN classifier score. Tournament win rate in this figure is the column-wise average of the pixel values in the heatmap. (Note that the classifier score at $i$=0 is lower than 4.0, which obscures the alignment between the rest of the curves when plotted on the same axis, so we omit it.) In the lower half of the figure: Figure~\ref{fig-monitor}b shows the same data but with matchups from far-apart iterations omitted, shown as grey pixels in Figure~\ref{fig-monitor}b-left. Figure~\ref{fig-monitor}b-right shows that skill rating continues to track the improvement of the model, even though some of the most informative battles (between early generators and later discriminators, in the top left) have been omitted. whereas the tournament win rate is no longer informative.}
  \label{fig-monitor}
\end{figure}

\subsection{Tournament win rate}
\label{winrate}

A tournament between a set of generators $\mathcal{G}$ and a set of discriminators $\mathcal{D}$ consists of a series of one-on-one matches between one generator and one discriminator.
We first describe a round-robin tournament, in which every pair in the
Cartesian product of the two sets participates in a match.

To determine the outcome of a match between discriminator $D$ and generator $G$, the discriminator $D$ judges two batches: one batch of samples from generator $G$, and one batch of real data. Every sample $x$ that is not judged correctly by the discriminator (e.g. $D(x) \ge 0.5$ for the generated data or $D(x) \le 0.5$ for the real data) counts as a win for the generator and is used to compute its \textit{win rate}. (Section~\ref{skillrating} elaborates on why we chose to include a batch of real data). A match win rate of 0.5 for $G$ means that $D$'s performance against $G$ is no better than random chance. The \textbf{tournament win rate} for generator $G$ is computed as its average win rate over all discriminators in $\mathcal{D}$. Tournament win rates are interpretable only within the context of the tournament they were produced from, and cannot be directly compared with those from other tournaments.

\subsection{Skill rating}
\label{skillrating}

Tournament win rate is simple to compute, and can be adequate for many purposes. However, its primary drawback is that each match carries equal weight. This can be undesirable if some of the matches contain redundant information, or if generators are not matched up against a balanced collection of both weak and strong discriminators. We introduce the idea of using a \textbf{skill rating} system
to summarize tournament outcomes in a way that takes into account the amount of new information each match provides. A skill rating system is a method for assigning a numerical skill to players in a player-vs-player game, given a win-loss record of games played. Higher ratings indicate higher player skill. Although skill rating systems are usually applied to symmetrical games, there is no restriction against the graph of matches being bipartite, so they can also be applied to asymmetrical games --- here, generators versus discriminators. Skill rating, like win rate, is comparable only in the context of a specific tournament.

Throughout this paper, we use the Glicko2 system \citep{Glickman2013}. To summarize briefly: each player's skill rating is represented as a Gaussian distribution, with a mean and standard deviation, representing the current state of the evidence about their ``true'' skill rating.
Because we use frozen snapshots of machine learning models, we disabled an irrelevant feature of Glicko2 that increases uncertainty about a human player's skill when they have not participated in a match for some time.

Both generators and discriminators are ``players'' in the game, and so although we only report the skill ratings of the generators in this work, the discriminators are also assigned a skill which is used in the overall computation: beating a ``stronger'' discriminator is evidence of higher generator skill. Including real data in the evaluation, as we describe in Section~\ref{winrate}, ensures that discriminators cannot be assigned the highest possible skill by outputting ``fake'' indiscriminately.

\section{Results}
\label{results}



\subsection{Within-trajectory tournaments to monitor GAN training}
\label{monitor}

One common use case of an evaluation method is to make sure the algorithm is successfully making progress as it trains.
We demonstrate that tournament outcomes from snapshots from a single learning trajectory can be used to validate that generators later in the experiment are indeed stronger than generators earlier in the experiment, even without access to discriminators from other experiments.

We run a tournament between 20 saved checkpoints of discriminators and generators from the same training run of a DCGAN \citep{Radford2015} trained on SVHN \citep{Netzer2011}.
(we use the identifier \dcgan{} to refer to this model).
We use an evaluation batch size of 64. We include slightly more checkpoints from earlier iterations. Figure~\ref{fig-monitor}(a) shows the raw tournament outcomes from the within-trajectory tournament, alongside the same tournament outcomes summarized using tournament win rate and skill rating (Sections~\ref{winrate} and~\ref{skillrating}), as well as SVHN classifier score \citep{Salimans2016} \footnote{SVHN classifier score here refers to the same procedure as Inception Score\citep{Salimans2016}, but using a pre-trained SVHN classifier rather than an ImageNet classifier.} and SVHN Fr\'echet distance\citep{Heusel2017}\footnote{Similarly, SVHN Fr\'echet distance is analogous to Fr\'echet Inception Distance (FID) \citep{Heusel2017}} computed from 10,000 samples, for comparison. We observe that tournament win rate and skill rating both provide a comparable measure of training progress to SVHN classifier score. 

\paragraph{Skill rating allows for running fewer battles.}
\label{omit}
Running all pairwise matchups between generators and discriminators might become prohibitively expensive as the number of checkpoints becomes large. Skill rating allows fewer matches to be run --- note that worldwide rankings in chess do not require every chess player in the world to compete with every other \citep{Glickman1995}. Figure~\ref{fig-monitor}(b) provides a proof-of-concept demonstration that skill rating allows battles to be omitted for efficiency. We run the same within-trajectory tournament as in Figure~\ref{fig-monitor}(a), but we omit matchups between checkpoints from far-apart iterations. Although tournament win rate performs poorly on this set of matches, skill rating has no difficulty rating the generators despite the imbalanced opponent pool. A full exploration of how match omission trades off against rating accuracy remains an open question for future work.  Note that our experiments throughout this paper use between 20-60 discriminators for the skill rating calculation, and so omitting any one discriminator does not substantially impact the outcomes, whereas in smaller tournaments a single discriminator's inclusion or omission may be more likely to have a large effect.

\paragraph{Tournament-based evaluation succeeds in unexplored domains.}
\label{apples}

Here we show preliminary evidence that tournament-based evaluation succeeds in domains that are poorly-represented by standard image embeddings.
Methods such as Inception Score and Fr\'echet Inception Distance have been widely adopted in the evaluation of generative models for natural images.
The main downside of these methods is that they depend on a good feature space, which may not be readily available for other kinds of data (see Section~\ref{context} for more context and comparison). As a proof-of-concept for an unexplored domain in which a standard feature space is not available, we evaluate a GAN trained on 70,000 hand-drawn images of apples from the QuickDraw \citep{Ha2017} dataset. Although we represent the drawings as images rather than strokes, they are not ``natural'' images (i.e. photographs of the physical world). We compare within-trajectory skill rating to evaluation methods that use a natural image embedding space from an unrelated dataset (SVHN).

\begin{figure}
  \centering
  \includegraphics[width=\linewidth]{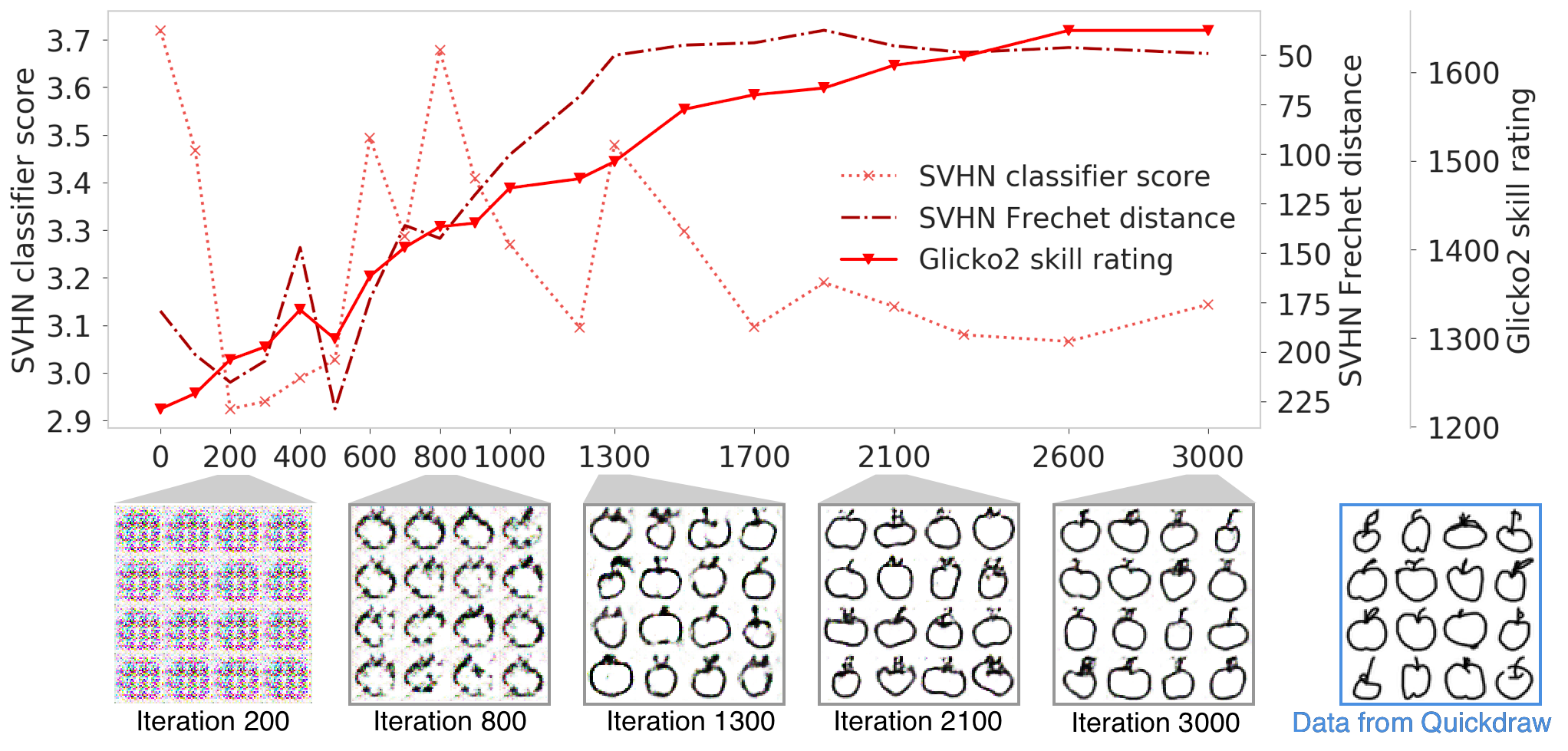}
  \caption{\textbf{Within-trajectory skill rating applied to drawings of apples.} We evaluate a DCGAN trained on drawings of apples from the QuickDraw dataset. From left to right, subjective sample quality improves with more iterations.
    SVHN Classifier score is a poor judge of quality for these samples, rating iteration 0 the highest, and providing choppy but broadly worsening ratings thereafter. SVHN Fr\'echet distance is a better fit, rating sample quality as steadily increasing until iteration 1300; however, it saturates at this point, whereas subjective sample quality continues increasing. (Note the inverted y-axis on the Fr\'echet distance plot, such that lower distance (better quality) is plotted higher on the plot). Within-trajectory skill rating continues improving beyond iteration 1300.
}
  \label{fig-apples}
\end{figure}

Figure~\ref{fig-apples} shows that, subjectively, sample quality increases consistently with more iterations. SVHN Classifier score is a poor judge of quality for these samples. Fr\'echet distance is a better fit, but saturates at iteration 1300 whereas sample quality continues improving. Of these three methods, within-trajectory skill rating is the best fit, providing preliminary evidence that skill rating can succeed in unexplored domains.

\subsection{Tournaments to compare GANs}
\label{compare}

Here we present the results of using a larger tournament to comparatively evaluate different trained GANs. We demonstrate that the resulting rankings anecdotally correlate with human perceptual preferences.

We construct a tournament from saved snapshots from six
SVHN GANs that differ slightly from one another, including different loss functions and architectures. We consider both helpful and harmful variations, to demonstrate that our evaluation method can tell which modifications are improvements and which are not. We want to emphasize that we are evaluating specific trained models, rather than general approaches. Our evaluation method is not intended to capture the best possible performance of an algorithm after all tuning has been completed. In order to discourage an interpretation that we are comparing general algorithms, we refer to these models by short identifiers, rather than a description of the training algorithm.
The details of the algorithms are presented in Appendix~\ref{svhn_architecture}.

Experiment \dcgan{} is an ordinary DCGAN, using the architecture, loss function, and hyperparameters from \citet{Gulrajani2017}, except with pixelnorm instead of batchnorm in the discriminator only, and noise added to the discriminator's input at training time. \footnote{We removed batchnorm from the discriminators out of concern that our results would be harder to interpret otherwise. When using discriminators to judge samples other than those they trained on, it's not clear which distribution should be used to set the batchnorm statistics. See Appendix~\ref{appendix-batchnorm} for results using batchnorm.}
Experiment \feature{} uses another different loss function.
Experiment \wgan{} uses the same architecture but a different loss function.
Experiments \conditional{} and \eleven{} use class-conditional architectures. The discriminators in these architectures require a label as auxiliary information, which is not available for arbitrary generated samples, and so only the generators from these models are eligible to participate in the tournament.
Experiment \pixelcnn{} is not a GAN, but rather an autoregressive model, which also participates only as a generator. We include only a single saved snapshot of \pixelcnn{}, not a full learning curve trajectory.


We include 20 saved checkpoints of discriminators and generators from each GAN experiment, a single snapshot of \pixelcnn{}, and a generator player that produces batches of real data as a benchmark. We sample slightly more checkpoints from earlier iterations, in order to provide more granular estimation in the region where performance is changing more rapidly. We run all pairwise matches betweeen these players. We do not correct for the fact that the discriminators from \conditional{}, \eleven{}, and \pixelcnn{} cannot participate in the tournament.

\begin{figure}
  \centering
  \includegraphics[width=1.0\linewidth]{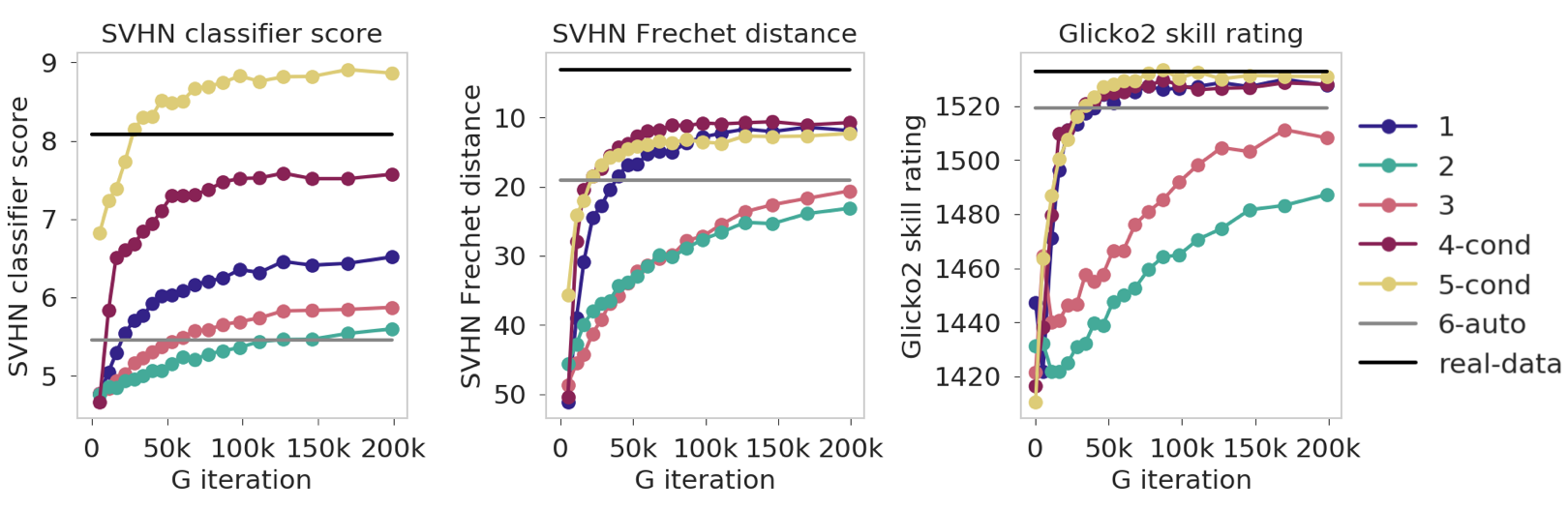}
  \caption{\textbf{Multiple-trajectory tournament outcomes.} We run a tournament containing SVHN generator and discriminator snapshots from models with different seeds,
	  hyperparameters, and architectures (described in Section~\ref{compare}). We evaluate them using SVHN classifier score (left), SVHN Fr\'echet distance (center), and our skill rating method (right; see Section~\ref{skillrating}). Each point represents the score of one iteration of one model. The overall trajectories show the improvement of each model with increasing training. Note the inverted y-axis on the Fr\'echet distance plot, such that lower distance (better quality) is plotted higher on the plot. The score of real data samples is shown as a black line. The score of \pixelcnn{} is evaluted from a single snapshot, rather than a full training curve, and is shown as a grey line. The learning curves produced by skill rating broadly agree with those produced by Fr\'echet distance, and disagree with classifier score only in the case of the conditional models \conditional{} and \eleven{} --- we speculate about this discrepancy in Section~\ref{compare}.}
  \label{fig-multi}
\end{figure}

\begin{figure}
  \centering
  \begin{subfigure}[b]{0.24\linewidth}
    \includegraphics[width=\linewidth]{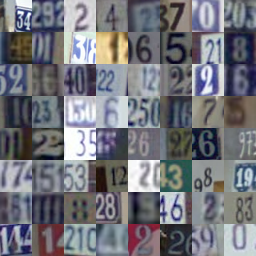}
    \caption{Real data samples\\\textbf{SR=1532}\\ CS=8.09 FD=3.03}
  \end{subfigure}
  \begin{subfigure}[b]{0.24\linewidth}
    \includegraphics[width=\linewidth]{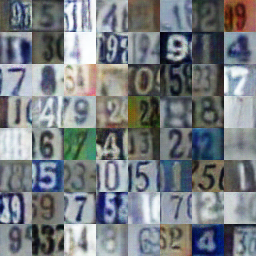}
    \caption{Experiment \eleven{}\\\textbf{SR=1530}\\ CS=8.86 FD=12.28}
  \end{subfigure}
  \begin{subfigure}[b]{0.24\linewidth}
    \includegraphics[width=\linewidth]{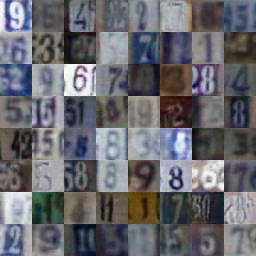}
    \caption{Experiment \conditional{}\\\textbf{SR=1528}\\ CS=7.57 FD=10.72}
  \end{subfigure}
  \begin{subfigure}[b]{0.24\linewidth}
    \includegraphics[width=\linewidth]{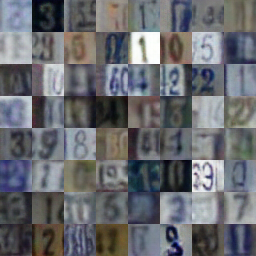}
    \caption{Experiment \dcgan{}\\\textbf{SR=1528}\\ CS=6.52 FD=11.86}
  \end{subfigure}
  \\
  \begin{subfigure}[b]{0.24\linewidth}
    \includegraphics[width=\linewidth]{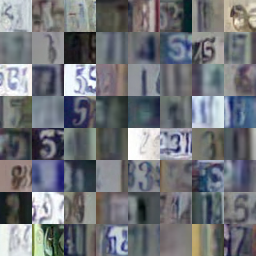}
    \caption{Experiment \pixelcnn{}\\\textbf{SR=1519}\\ CS=5.46 FD=19.10}
  \end{subfigure}
  \begin{subfigure}[b]{0.24\linewidth}
    \includegraphics[width=\linewidth]{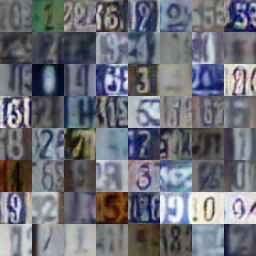}
    \caption{Experiment \wgan{}\\\textbf{SR=1508}\\ CS=5.87 FD=20.61}
  \end{subfigure}
  \begin{subfigure}[b]{0.24\linewidth}
    \includegraphics[width=\linewidth]{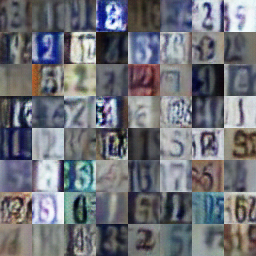}
    \caption{Experiment \feature{}\\\textbf{SR=1487}\\ CS=5.60 FD=23.13}
  \end{subfigure}
  \begin{subfigure}[b]{0.24\linewidth}
    \phantom{Phantom box testing x}
  \end{subfigure}
  \caption{\textbf{Samples from fully-trained generative models}. From each trained model, we show 64 samples (from iteration 200,000 of the GANs and epoch 106 of \pixelcnn{}), along with real data for comparison. Under each set of samples, we list the Glicko2 skill rating (SR), SVHN classifier score (CS), and SVHN Fr\'echet distance (FD) of the model. Our skill rating system ranks experiment \eleven{} as being slightly worse than real data and slightly better than runner-ups \conditional{} and \dcgan{}, whereas classifier score ranks \eleven{} better than real data, and Fr\'echet distance ranks \eleven{} worse than both \conditional{} and \dcgan{}. Our system's rankings agree with Fr\'echet distance in all other cases.}
  \label{fig-samples}
\end{figure}

Figure~\ref{fig-multi} shows skill rating, classifier score, and Fr\'echet distance trajectories from the tournament of all the above players. Figure~\ref{fig-samples} shows samples and scores from the final trained models. Win rate heatmaps (similar to Figure~\ref{fig-monitor}a-left) can be found in Appendix~\ref{appendix-winrate}. We note first a difference in the ranking of conditional architectures. Our method ranks \eleven{} as the highest-quality model, but not as high-quality as real data. Classifier score ranks \eleven{} even higher than real data. Fr\'echet distance ranks \eleven{} as lower-quality than both \conditional{} and \dcgan{}. In our anecdotal judgment, we believe that our method's ranking agrees most with our subjective visual assessment of sample quality.

Secondly, we consider the ranking of \pixelcnn{}. These samples were not produced by a GAN, and have different strengths and weaknesses than GAN samples. We were interested to know whether GAN discriminators can correctly evaluate samples that were produced by an entirely different generative approach. Our method agrees with Fr\'echet distance in the ranking of these samples, whereas classifier score ranks them beneath \feature{} and \wgan{}. In our anecdotal judgment, either of these rankings could be considered correct: \pixelcnn{} is more likely to produce blurry samples, whereas \feature{} and \wgan{} are more likely to produce wobbly samples; all produce a similar proportion of clear, recognizable samples. We conclude that our method assigns unfamiliar samples a rank ordering that broadly agrees with subjective human judgment in this case.

Finally, we note that our method has ranked real data quite closely to the top-ranking models. This compression of ratings was not seen in previous experiments (see Appendix~\ref{appendix-batchnorm}). Our current speculation is that the discriminators here are less discerning overall than the discriminators from our earlier experiments, and so are more fooled by the best generated samples. We acknowledge that, depending on the tournament population, discriminators may not accurately judge just \emph{how} much better the real data is than the generated samples, even though the final ranking is correct here.

\captionsetup[subfigure]{width=1.0\textwidth}
\captionsetup[subfigure]{width=0.9\textwidth}

\subsection{Toy problem: evaluating near-perfect generators}
\label{chekhov}
\begin{figure}
  \centering
  \begin{subfigure}[b]{0.28\linewidth}
    \includegraphics[width=\linewidth]{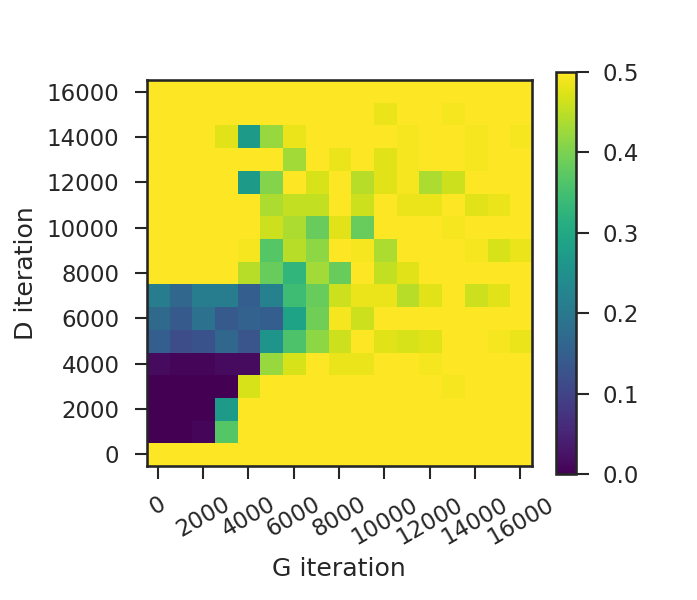}
    \caption{Ordinary GAN generator vs. own discriminator}
    \label{fig-chekhov-self-heatmap}
  \end{subfigure}
  \begin{subfigure}[b]{0.28\linewidth}
    \includegraphics[width=\linewidth]{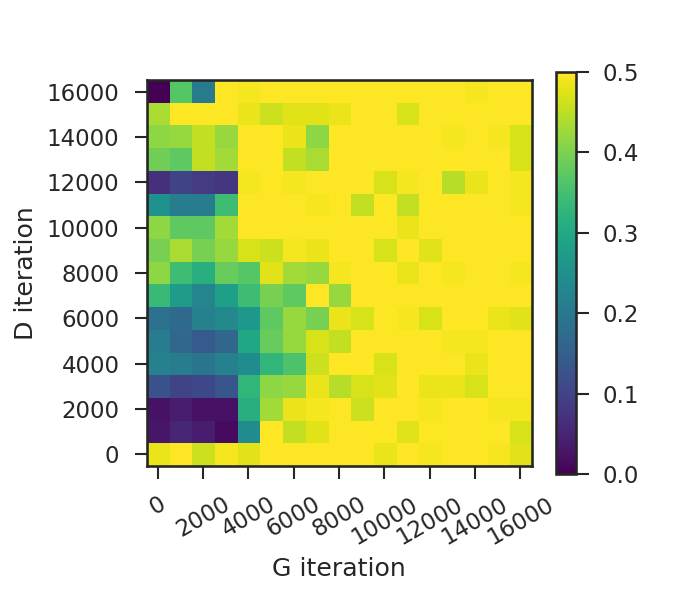}
    \caption{Ordinary generator vs. Chekhov discriminator}
    \label{fig-chekhov-chekhov-heatmap}
  \end{subfigure}
  \begin{subfigure}[b]{0.39\linewidth}
    \includegraphics[width=\linewidth]{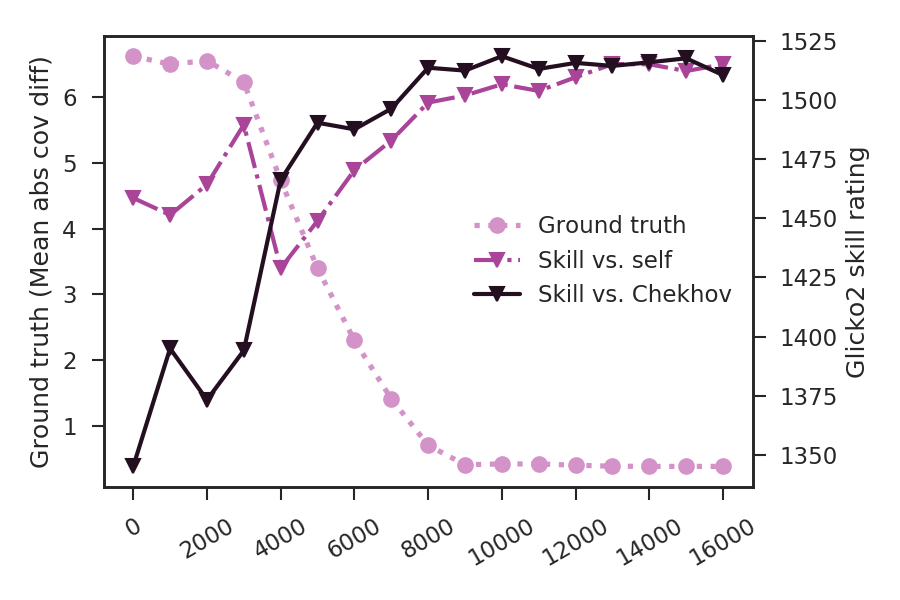}
    \caption{Skill rating for ordinary-vs-self and ordinary-vs-Chekhov, compared to ``ground truth'' performance}
    \label{fig-chekhov-skillrating}
  \end{subfigure}
  \caption{\textbf{Evaluating a near-perfect generator on a toy problem}. We train an ordinary GAN to model a Gaussian distribution with a full covariance matrix. Generators from iteration 8000 onwards have mastered this task. Discriminators from iteration 8000 onwards no longer produce useful judgments (Figure~\ref{fig-chekhov-self-heatmap}). Chekhov GAN discriminators beyond iteration 8000 retain their ability to judge past generators' samples (Figure~\ref{fig-chekhov-chekhov-heatmap}). Figure~\ref{fig-chekhov-skillrating} compares skill ratings from these discriminators with the ground truth performance of the ordinary generator, measured as the mean absolute difference between the generator's estimated covariance matrix and that of the data. Skill ratings against the Chekhov discriminator were a better fit to the ground truth than those from within-trajectory matches.}
  \label{fig-chekhov}
\end{figure}

For complex real-world datasets, generative models do not currently succeed at
learning the target data distribution perfectly. However, for simpler datasets,
it is possible for the generator to attain near-perfect performance, in which case the discriminator's output from that point onward becomes effectively unconstrained. To verify that tournament-based evaluation can be applied even in such settings, we experimented with a toy task that is easy for the generator to solve: modeling a Gaussian distribution with a full covariance matrix. In this case, we found that once the generator has mastered the task, discriminators from that iteration onwards no longer produce useful judgments (Figure~\ref{fig-chekhov-self-heatmap}). We resolved this problem by evaluating the generator from the ordinary model against the discriminator from a Chekhov GAN \citep{Grnarova2017} rather than against its own discriminator. Chekhov GANs train each player against several past versions of their opponent (we use 10 past opponents, selected with reservoir sampling). We found empirically that Chekhov GAN discriminators retained their ability to judge past generators' samples even after the generator they trained with achieved nearly-perfect performance (Figure~\ref{fig-chekhov-chekhov-heatmap}). The resulting skill ratings from matches against the Chekhov GAN discriminator were a better fit to the ground truth performance of the generator than those from the within-trajectory matches (Figure~\ref{fig-chekhov-skillrating}).

This experiment shows that difficulties applying skill rating can arise in some cases, and so blind trust in the method is not warranted. However, our experience in this case also suggests that difficulties can be resolved with attention to the pattern of match outcomes. If specific anomalies are observed, they can be remedied by thoughtfully selecting discriminators designed to address the problem.
In this case, no modification was required to the discriminators used at training-time; all that was needed was to include discriminators in the evaluation set that were designed to avoid catastrophic forgetting. Full details of the toy task and the GAN architectures are specified in Appendix~\ref{toy_task_architecture}.

\section{Future work}
\label{future}

One direction in which to extend this work is in the specific format of the tournament.
In this paper, games are played over single samples, so generators that suffer from low diversity can perform
well in these tournaments, but this could be resolved with tournaments that involve games played at the batch level.
We also use a binary threshold, counting a ``win'' for the generator if the discriminator rates a generated
sample as real with $D(x) \ge 0.5$, but we could experiment with alternate ways of using the discriminator's output.

We note that the discriminators in these tournaments are designed to rate a given sample as ``real'' or ``fake'' depending on which distribution it is \emph{comparatively more similar} to, even if it is highly dissimilar to both distributions. There is no particular constraint that would necessarily lead previously-unseen data to be labeled as ``fake''. Future work might investigate using moment-matching discriminators for tournament-based evaluation, after configured them to use ``distance from \emph{real data} in feature-space'' when making judgments. Asymmetrically privileging the real data distribution at evaluation-time could help the discriminators reject unfamiliar generated samples more effectively. In Appendix~\ref{eoe} we show some exploratory analyses of skill rating's performance on distorted real samples: one might expect distance-based discriminators to be more likely to give monotonically lower ratings to progressively greater levels of distortion than the discriminators used here.

We show in Section~\ref{omit} that it is possible to skill rate all $n$ players in a tournament without needing to run $n^2$ matches,
but we do not yet undertake a full exploration of how to determine which matches may be omitted.
In Section~\ref{context} we note that reproducing scores requires reproducing the population of models used in the tournament: specifically, rating a new model against published skill ratings and models from an $N$-model tournament could require as many as $2N$ more matches to be run ($D_1, D_2,... D_N$ vs $G_{N+1}$ and $D_{N+1}$ vs $G_1, G_2, ... G_N$; the existing matches do not need to be re-run, even if the match outcomes have been lost, because the numerical skill ratings contain the necessary information). However, this number is likely to be smaller in practice, just as a new chess player need not play every other chess player in the world to get an accurate rating; this remains to be fully demonstrated. In general, a more rigorous comparison of the computational complexity of our method as compared with others would be useful for determining the strengths and weaknesses of different evaluation methods.

Finally, we note that human judges are eligible to play as discriminators, and could participate to receive a skill rating. This might allow human perceptual evaluation to be incorporated into the evaluation of generative models in a more nuanced fashion, by taking into account the variation in judgment among human raters (See Section~\ref{context}).


As we mention in Section~\ref{context}, we provide empirical evidence that GAN discriminators can successfully judge samples from generators other than the one they trained against, but a full exploration of when this behavior can be expected remains an open question.


\subsubsection*{Acknowledgments}
Thanks to \citet{Gulrajani2017, Salimans2016}, and Joel Shor and Sergio Guadarrama of \texttt{tensorflow.contrib.gan} for releasing code which we were able to build upon in this work. Thanks to Dandelion Man\'e for contributions to an early prototype of this work.


\clearpage
\clearpage

\bibliography{paper}

\begin{thebibliography}{28}
\providecommand{\natexlab}[1]{#1}
\providecommand{\url}[1]{\texttt{#1}}
\expandafter\ifx\csname urlstyle\endcsname\relax
  \providecommand{\doi}[1]{doi: #1}\else
  \providecommand{\doi}{doi: \begingroup \urlstyle{rm}\Url}\fi

\bibitem[Arora et~al.(2018)Arora, Risteski, and Zhang]{arora2018do}
Sanjeev Arora, Andrej Risteski, and Yi~Zhang.
\newblock Do {GAN}s learn the distribution? some theory and empirics.
\newblock In \emph{International Conference on Learning Representations}, 2018.
\newblock URL \url{https://openreview.net/forum?id=BJehNfW0-}.

\bibitem[Borji(2018)]{Borji2018}
A.~Borji.
\newblock Pros and cons of gan evaluation measure.
\newblock arXiv:1802.03446, Feb 2018.
\newblock URL \url{https://arxiv.org/pdf/1802.03446}.

\bibitem[Breuleux et~al.(2010)Breuleux, Bengio, and
  Vincent]{Breuleux+al-TR-2010}
Olivier Breuleux, Yoshua Bengio, and Pascal Vincent.
\newblock Unlearning for better mixing.
\newblock Technical Report 1349, Universit{\'{e}} de Montr{\'{e}}al/DIRO, 2010.

\bibitem[Denton et~al.(2015)Denton, Chintala, Szlam, and
  Fergus]{denton2015deep}
Emily Denton, Soumith Chintala, Arthur Szlam, and Rob Fergus.
\newblock Deep generative image models using a {Laplacian} pyramid of
  adversarial networks.
\newblock \emph{NIPS}, 2015.

\bibitem[Durugkar et~al.(2017)Durugkar, Gemp, and Mahadevan]{Durugkar2017}
Ishan Durugkar, Ian Gemp, and Sridhar Mahadevan.
\newblock Generative multi-adversarial networks.
\newblock \emph{ICLR}, 2017.
\newblock URL \url{https://arxiv.org/pdf/1611.01673.pdf}.
\newblock arXiv:1611.01673.

\bibitem[Elo(1978)]{Elo1978}
A.~Elo.
\newblock \emph{The rating of chessplayers, past and present}.
\newblock Arco Pub., New York, 1978.
\newblock ISBN 0668047216 9780668047210.

\bibitem[Glickman(1995)]{Glickman1995}
M.~Glickman.
\newblock A comprehensive guide to chess ratings.
\newblock 1995.
\newblock URL \url{http://www.glicko.net/research/acjpaper.pdf}.

\bibitem[Glickman(2013)]{Glickman2013}
M.~Glickman.
\newblock Example of the glicko-2 system.
\newblock Nov 2013.
\newblock URL \url{http://www.glicko.net/glicko/glicko2.pdf}.

\bibitem[Goodfellow(2014)]{Goodfellow-ICLR2015}
Ian~J. Goodfellow.
\newblock On distinguishability criteria for estimating generative models.
\newblock In \emph{International Conference on Learning Representations,
  Workshops Track}, 2014.

\bibitem[Goodfellow et~al.(2014)Goodfellow, Pouget-Abadie, Mirza, Xu,
  Warde-Farley, Ozair, Courville, and Bengio]{Goodfellow-et-al-NIPS2014-small}
Ian~J. Goodfellow, Jean Pouget-Abadie, Mehdi Mirza, Bing Xu, David
  Warde-Farley, Sherjil Ozair, Aaron Courville, and Yoshua Bengio.
\newblock Generative adversarial networks.
\newblock In \emph{NIPS'2014}, 2014.

\bibitem[Grnarova et~al.(2017)Grnarova, Levy, Lucchi, Hofmann, and
  Krause]{Grnarova2017}
P.~Grnarova, K.~Levy, A.~Lucchi, T.~Hofmann, and A.~Krause.
\newblock An online learning approach to generative adversarial networks.
\newblock arXiv:1706.03269, Jun 2017.
\newblock URL \url{https://arxiv.org/abs/1706.03269}.

\bibitem[Gulrajani(2017)]{Gulrajani2017}
I.~Gulrajani.
\newblock Improved training of wasserstein gans.
\newblock \url{https://github.com/igul222/improved_wgan_training}, 2017.
\newblock fa66c574a54c4916d27c55441d33753dcc78f6bc.

\bibitem[Gutmann and Hyvarinen(2010)]{Gutmann+Hyvarinen-2010}
M.~Gutmann and A.~Hyvarinen.
\newblock Noise-contrastive estimation: A new estimation principle for
  unnormalized statistical models.
\newblock In \emph{aistats10}, 2010.

\bibitem[Ha and Eck(2017)]{Ha2017}
D.~Ha and D.~Eck.
\newblock A neural representation of sketch drawings.
\newblock arXiv:1704.03477, Apr 2017.
\newblock URL \url{https://arxiv.org/abs/1704.03477}.

\bibitem[Herbrich et~al.(2007)Herbrich, Minka, and Graepel]{Herbrich2007}
Ralf Herbrich, Tom Minka, and Thore Graepel.
\newblock Trueskill: a bayesian skill rating system.
\newblock In \emph{Advances in neural information processing systems}, pages
  569--576, 2007.

\bibitem[Heusel et~al.(2017)Heusel, Ramsauer, Unterthiner, Nessler, and
  Hochreiter]{Heusel2017}
M.~Heusel, H.~Ramsauer, T.~Unterthiner, B.~Nessler, and S.~Hochreiter.
\newblock Gans trained by a two time-scale update rule converge to a local nash
  equilibrium.
\newblock In \emph{Advances in Neural Information Processing Systems 30}. 2017.
\newblock URL
  \url{http://papers.nips.cc/paper/7240-gans-trained-by-a-two-time-scale-update-rule-converge-to-a-local-nash-equilibrium.pdf}.

\bibitem[Im et~al.(2016)Im, Kim, Jiang, and Memisevic]{im2016generating}
Daniel~Jiwoong Im, Chris~Dongjoo Kim, Hui Jiang, and Roland Memisevic.
\newblock Generating images with recurrent adversarial networks.
\newblock \emph{arXiv preprint arXiv:1602.05110}, 2016.

\bibitem[Karras et~al.(2017)Karras, Aila, Laine, and Lehtinen]{Karras2017}
Tero Karras, Timo Aila, Samuli Laine, and Jaakko Lehtinen.
\newblock Progressive growing of gans for improved quality, stability, and
  variation.
\newblock \emph{CoRR}, abs/1710.10196, 2017.
\newblock URL \url{http://arxiv.org/abs/1710.10196}.

\bibitem[Netzer et~al.(2011)Netzer, Wang, Coates, Bissacco, Wu, and
  Ng]{Netzer2011}
Y.~Netzer, T.~Wang, A.~Coates, A.~Bissacco, B.~Wu, and A.~Ng.
\newblock Reading digits in natural images with unsupervised feature learning.
\newblock In \emph{NIPS Workshop on Deep Learning and Unsupervised Feature
  Learning}. 2011.
\newblock URL \url{http://ufldl.stanford.edu/housenumbers/}.

\bibitem[OpenAI(2017)]{Openai2017}
OpenAI.
\newblock More on dota 2.
\newblock https://blog.openai.com/more-on-dota-2/, Aug 2017.

\bibitem[Radford et~al.(2015)Radford, Metz, and Chintala]{Radford2015}
A.~Radford, L.~Metz, and S.~Chintala.
\newblock Unsupervised representation learning with deep convolutional
  generative adversarial networks.
\newblock arXiv:1511.06434, Nov 2015.
\newblock URL \url{https://arxiv.org/abs/1511.06434}.

\bibitem[Salimans et~al.(2016)Salimans, Goodfellow, Zaremba, Cheung, Radford,
  and Chen]{Salimans2016}
T.~Salimans, I.~Goodfellow, W.~Zaremba, V.~Cheung, A.~Radford, and X.~Chen.
\newblock Improved techniques for training gans.
\newblock In \emph{Advances in Neural Information Processing Systems 29}. 2016.
\newblock URL
  \url{http://papers.nips.cc/paper/6125-improved-techniques-for-training-gans.pdf}.

\bibitem[Salimans et~al.(2017)Salimans, Karpathy, Chen, and
  Kingma]{Salimans2017}
T.~Salimans, A.~Karpathy, X.~Chen, and D.~Kingma.
\newblock Pixelcnn++: Improving the pixelcnn with discretized logistic mixture
  likelihood and other modifications.
\newblock arXiv:1701.05517, Jan 2017.
\newblock URL \url{https://arxiv.org/abs/1701.05517}.

\bibitem[Santurkar et~al.(2018)Santurkar, Schmidt, and Madry]{santurkar2018a}
Shibani Santurkar, Ludwig Schmidt, and Aleksander Madry.
\newblock A classification-based perspective on {GAN} distributions, 2018.
\newblock URL \url{https://openreview.net/forum?id=S1FQEfZA-}.

\bibitem[Silberman et~al.(2015)Silberman, Milland, LaPlante, Ross, and
  Irani]{Silberman2015}
S.~Silberman, K.~Milland, R.~LaPlante, J.~Ross, and L.~Irani.
\newblock Stop citing ross et al. 2010, ``who are the crowdworkers''?
\newblock
  \url{https://medium.com/@silberman/stop-citing-ross-et-al-2010-who-are-the-crowdworkers-b3b9b1e8d300},
  Mar 2015.

\bibitem[Silver et~al.(2016)Silver, Huang, Maddison, Guez, Sifre, Van
  Den~Driessche, Schrittwieser, Antonoglou, Panneershelvam, Lanctot,
  et~al.]{Silver2016}
David Silver, Aja Huang, Chris~J Maddison, Arthur Guez, Laurent Sifre, George
  Van Den~Driessche, Julian Schrittwieser, Ioannis Antonoglou, Veda
  Panneershelvam, Marc Lanctot, et~al.
\newblock Mastering the game of go with deep neural networks and tree search.
\newblock \emph{nature}, 529\penalty0 (7587):\penalty0 484, 2016.

\bibitem[{Szegedy} et~al.(2015){Szegedy}, {Vanhoucke}, {Ioffe}, {Shlens}, and
  {Wojna}]{Szegedy-et-al-2015}
C.~{Szegedy}, V.~{Vanhoucke}, S.~{Ioffe}, J.~{Shlens}, and Z.~{Wojna}.
\newblock {Rethinking the Inception Architecture for Computer Vision}.
\newblock \emph{ArXiv e-prints}, December 2015.

\bibitem[Theis et~al.(2016)Theis, van~den Oord, and Bethge]{Theis2015d}
L.~Theis, A.~van~den Oord, and M.~Bethge.
\newblock A note on the evaluation of generative models.
\newblock arXiv:1511.01844, 2016.
\newblock URL \url{http://arxiv.org/abs/1511.01844}.

\end{thebibliography}
\bibliographystyle{plainnat}

\clearpage
\clearpage

\appendix

\section{Appendix: Raw win rate matrices}
\label{appendix-winrate}

\begin{figure}
  \centering
  \begin{subfigure}[b]{0.3\linewidth}
    \includegraphics[width=\linewidth]{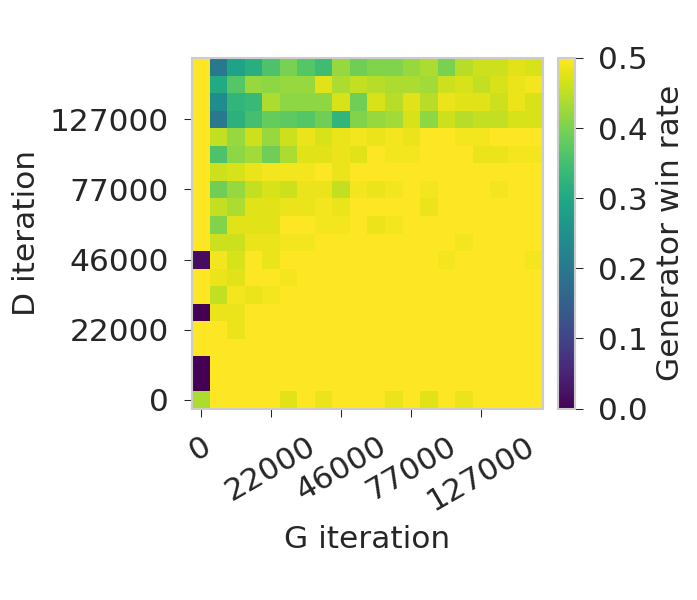}
    \vspace*{-8mm}
    \caption{D=\dcgan{} vs G=\dcgan{}}
  \end{subfigure}
  \begin{subfigure}[b]{0.3\linewidth}
    \includegraphics[width=\linewidth]{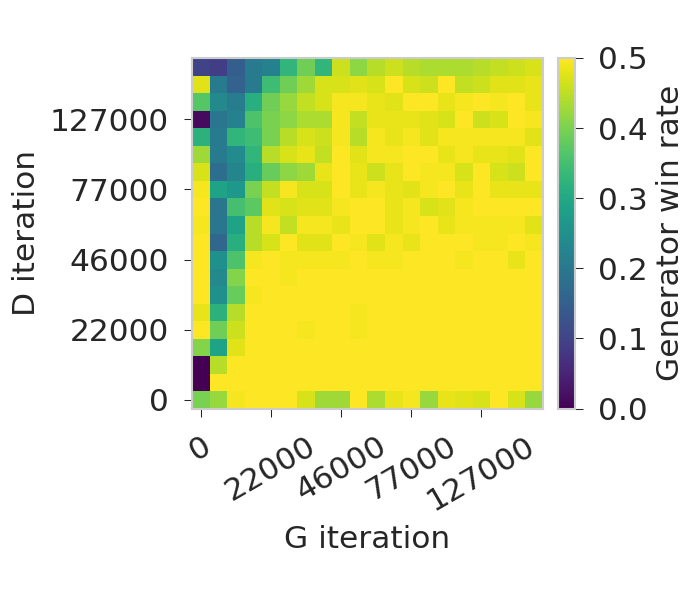}
    \vspace*{-8mm}
    \caption{D=\feature{} vs G=\dcgan{}}
  \end{subfigure}
  \begin{subfigure}[b]{0.3\linewidth}
    \includegraphics[width=\linewidth]{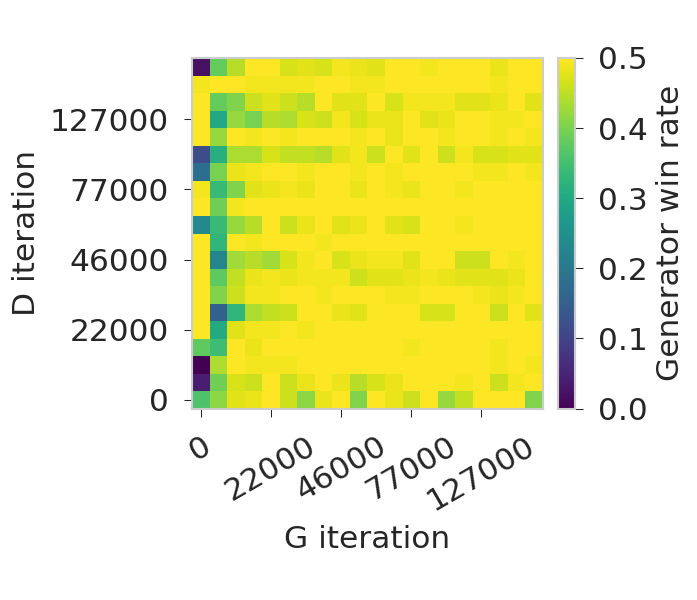}
    \vspace*{-8mm}
    \caption{D=\wgan{} vs G=\dcgan{}}
  \end{subfigure}
  \\
  \begin{subfigure}[b]{0.3\linewidth}
    \includegraphics[width=\linewidth]{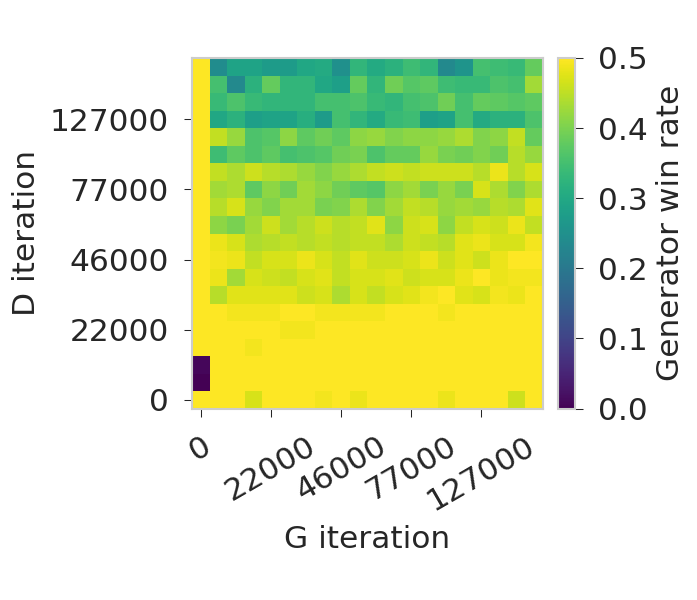}
    \vspace*{-8mm}
    \caption{D=\dcgan{} vs G=\feature{}}
  \end{subfigure}
  \begin{subfigure}[b]{0.3\linewidth}
    \includegraphics[width=\linewidth]{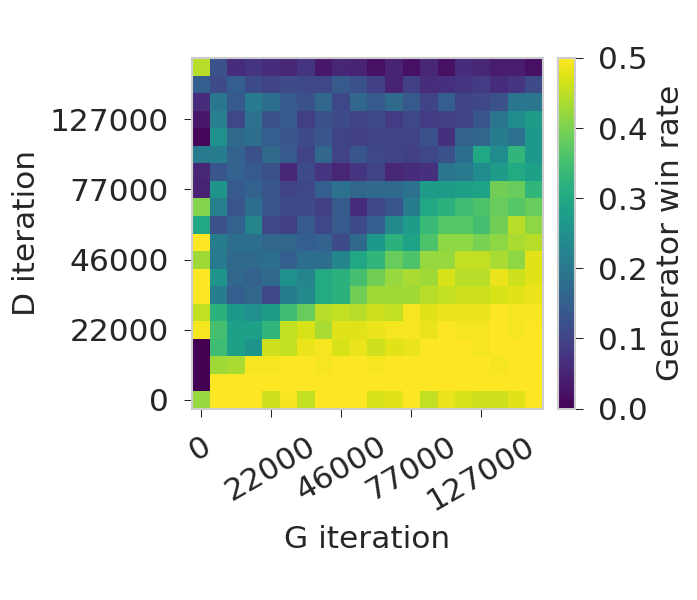}
    \vspace*{-8mm}
    \caption{D=\feature{} vs G=\feature{}}
  \end{subfigure}
  \begin{subfigure}[b]{0.3\linewidth}
    \includegraphics[width=\linewidth]{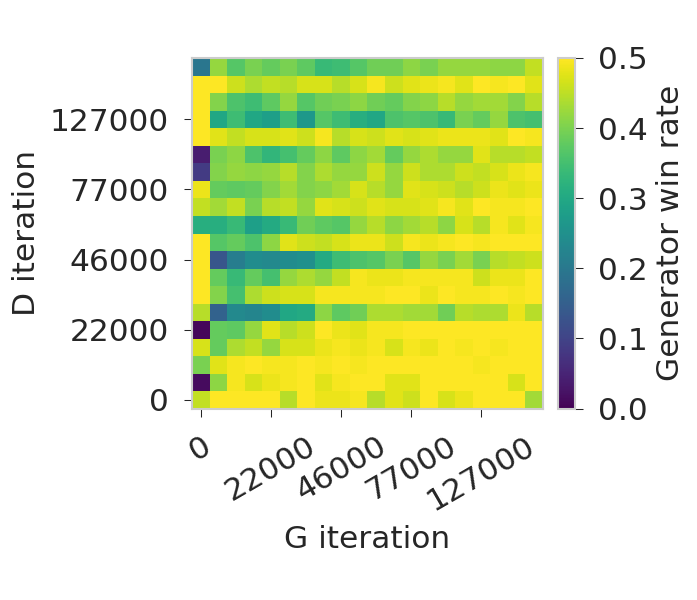}
    \vspace*{-8mm}
    \caption{D=\wgan{} vs G=\feature{}}
  \end{subfigure}
  \\
  \begin{subfigure}[b]{0.3\linewidth}
    \includegraphics[width=\linewidth]{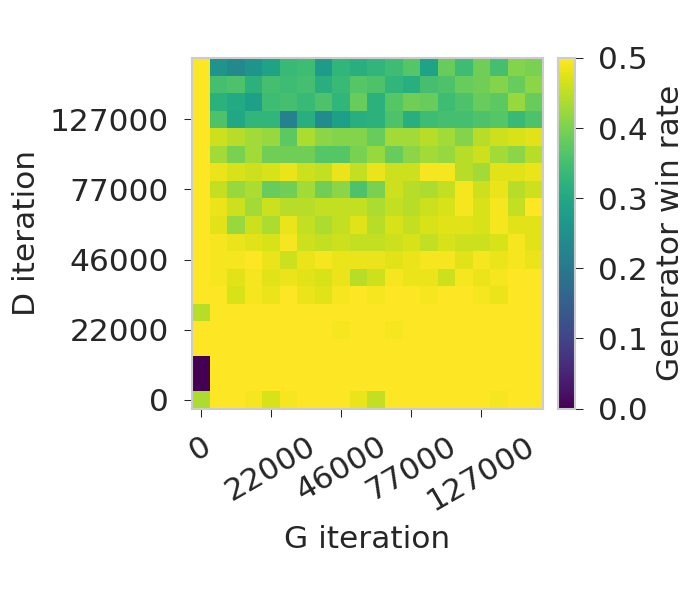}
    \vspace*{-8mm}
    \caption{D=\dcgan{} vs G=\wgan{}}
  \end{subfigure}
  \begin{subfigure}[b]{0.3\linewidth}
    \includegraphics[width=\linewidth]{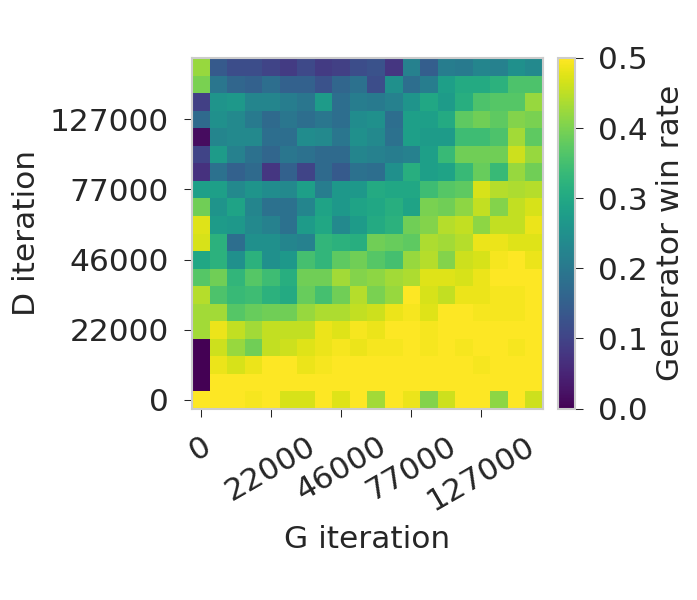}
    \vspace*{-8mm}
    \caption{D=\feature{} vs G=\wgan{}}
  \end{subfigure}
  \begin{subfigure}[b]{0.3\linewidth}
    \includegraphics[width=\linewidth]{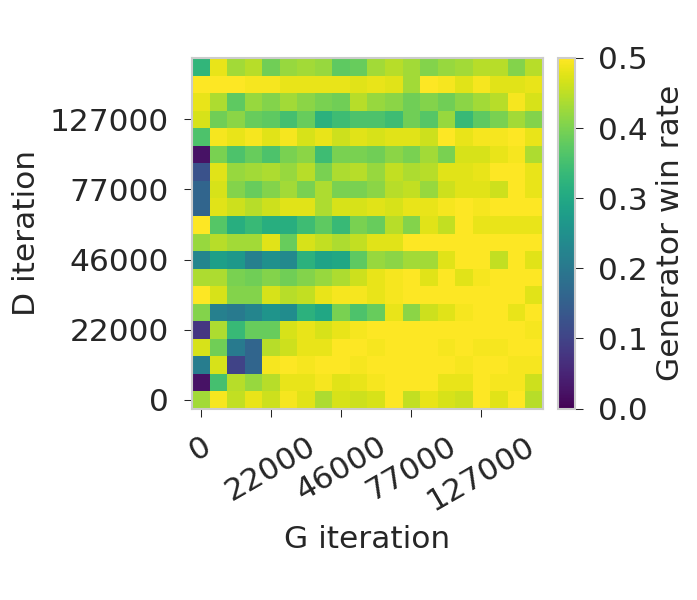}
    \vspace*{-8mm}
    \caption{D=\wgan{} vs G=\wgan{}}
  \end{subfigure}
  \\
  \begin{subfigure}[b]{0.3\linewidth}
    \includegraphics[width=\linewidth]{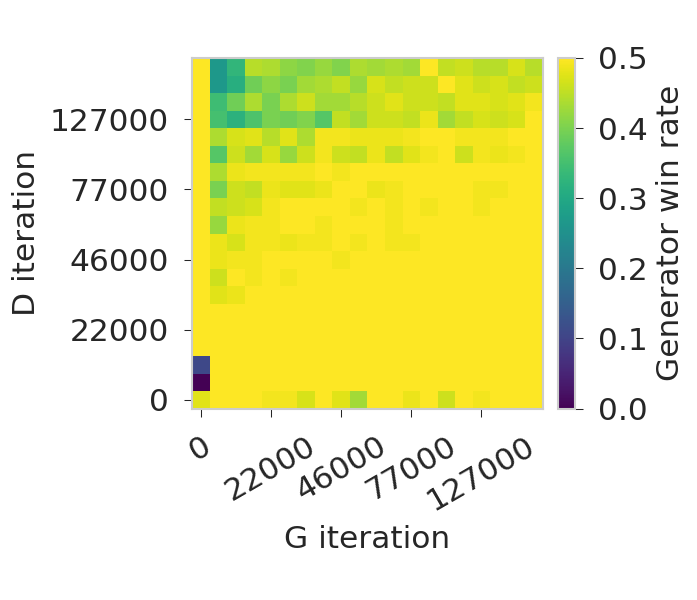}
    \vspace*{-8mm}
    \caption{D=\dcgan{} vs G=\conditional{}}
  \end{subfigure}
  \begin{subfigure}[b]{0.3\linewidth}
    \includegraphics[width=\linewidth]{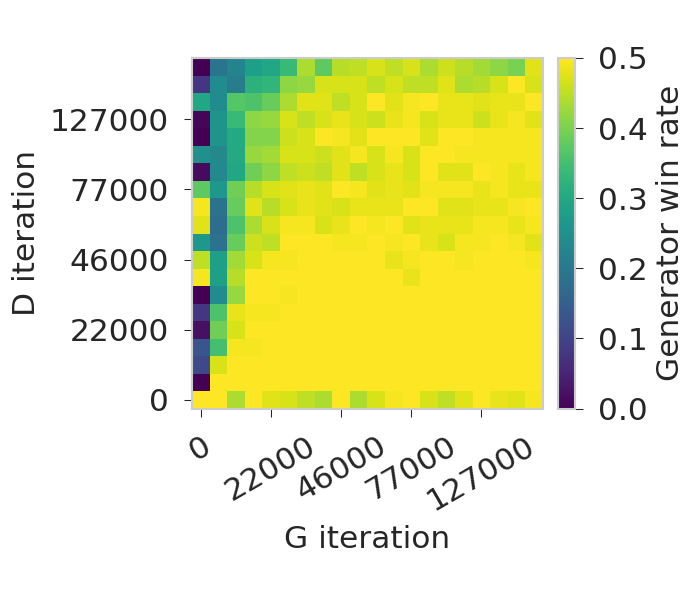}
    \vspace*{-8mm}
    \caption{D=\feature{} vs G=\conditional{}}
  \end{subfigure}
  \begin{subfigure}[b]{0.3\linewidth}
    \includegraphics[width=\linewidth]{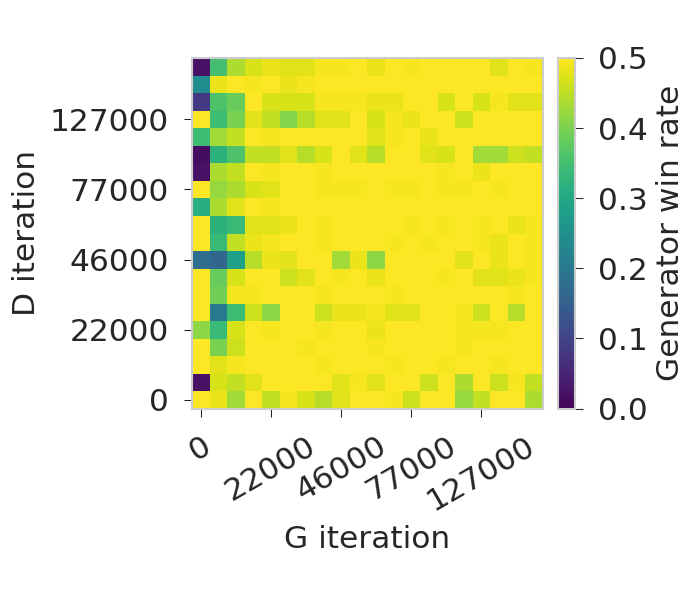}
    \vspace*{-8mm}
    \caption{D=\wgan{} vs G=\conditional{}}
  \end{subfigure}
  \\
  \begin{subfigure}[b]{0.3\linewidth}
    \includegraphics[width=\linewidth]{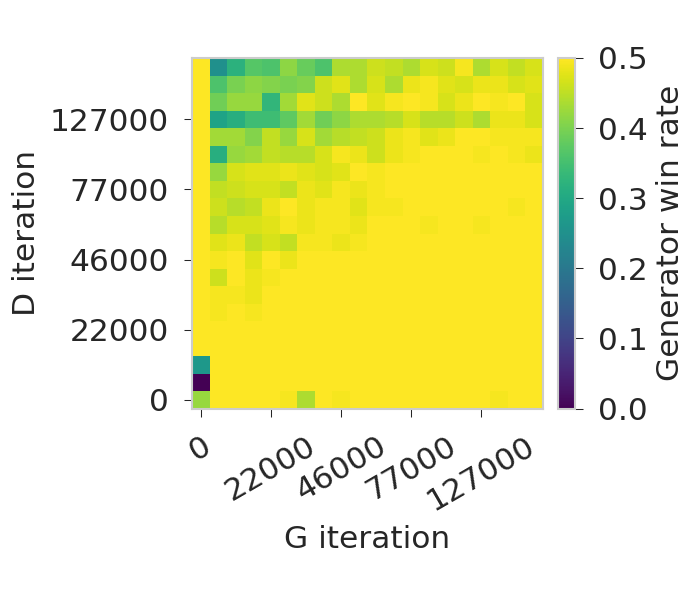}
    \vspace*{-8mm}
    \caption{D=\dcgan{} vs G=\eleven{}}
  \end{subfigure}
  \begin{subfigure}[b]{0.3\linewidth}
    \includegraphics[width=\linewidth]{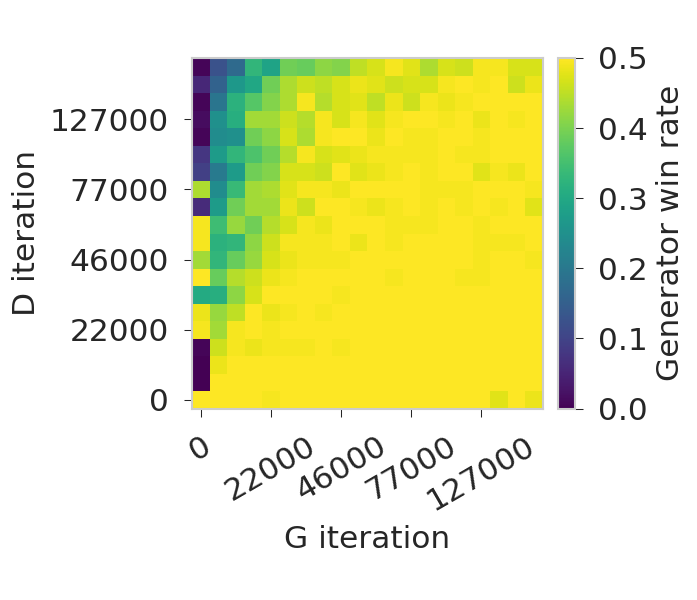}
    \vspace*{-8mm}
    \caption{D=\feature{} vs G=\eleven{}}
  \end{subfigure}
  \begin{subfigure}[b]{0.3\linewidth}
    \includegraphics[width=\linewidth]{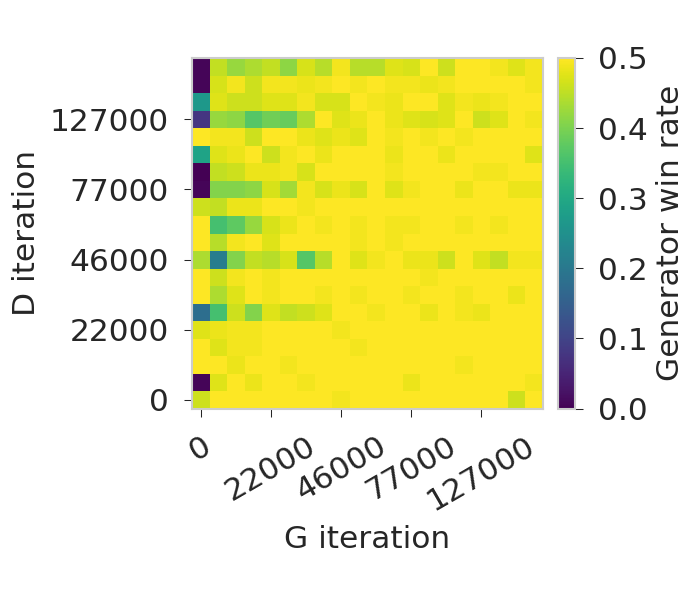}
    \vspace*{-8mm}
    \caption{D=\wgan{} vs G=\eleven{}}
  \end{subfigure}
  \\
  \caption{\textbf{Tournament heatmaps}. Win rate heatmaps from the tournament in Section~\ref{compare} are shown here for each individual generator and discriminator matchup. Each column contains results from one of the three GAN discriminators in the tournament: \dcgan{}, \wgan{}, and \feature{}. Each row contains results from one of the five GAN generators in the tournament (including additionally \conditional{} and \eleven{}).}
  \label{fig-appendix-winrate}
\end{figure}

In Figure~\ref{fig-appendix-winrate} we show full win rate heatmaps (similar to Figure~\ref{fig-monitor}a-left) for the tournament described in Section~\ref{compare}.

\captionsetup[subfigure]{width=1.0\textwidth}
\captionsetup[subfigure]{width=0.9\textwidth}

\section{Appendix: Evaluation on distorted samples}
\label{eoe}
\begin{figure}
  \centering
  \includegraphics[width=1.0\linewidth]{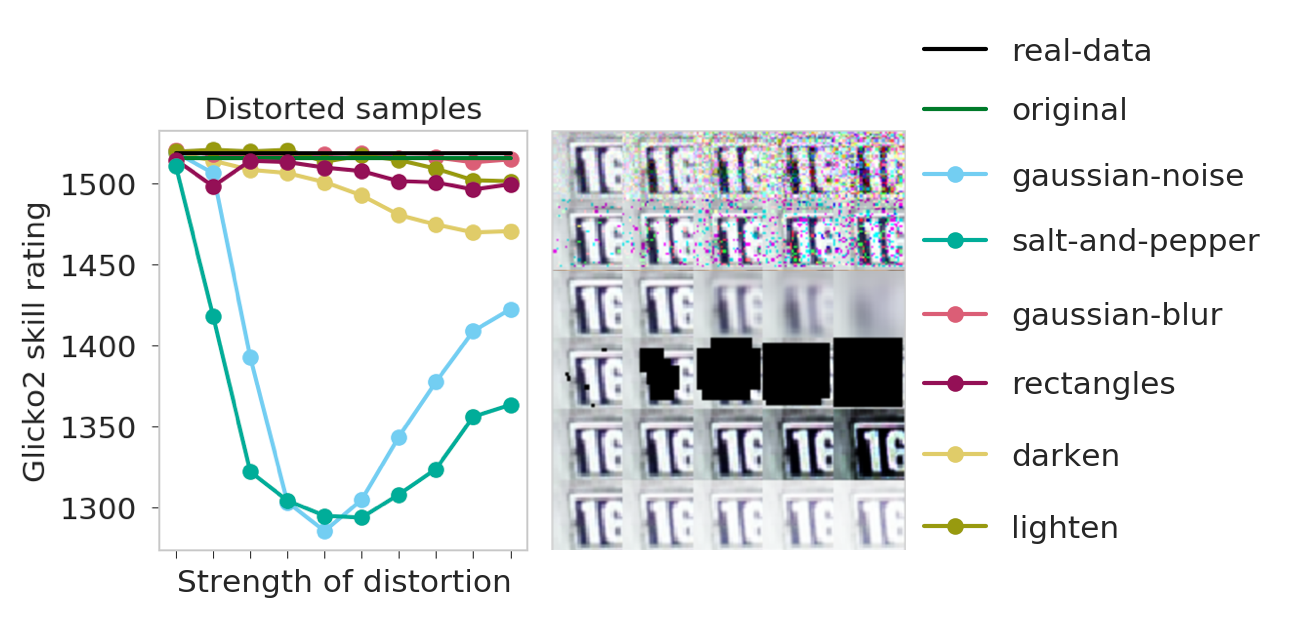}
  \caption{\textbf{Evaluation of skill rating on distorted samples.} We use the discriminators from experiments \dcgan{}, \wgan{}, and \feature{} to skill rate SVHN samples which have been distorted using six different distortion procedures, at nine different strengths/severities of distortion. At center, each row of samples shows distortion strength level 1, 3, 5, 7, and 9 of a given transform, with the name of the transform and its corresponding graph legend at the right. ``Real data'' is the batch of data which has been transformed, and ``original'' is another different batch of real data. The skill rating procedure is not very sensitive to gaussian blur. It is reasonably sensitive to lightening the image, and yet more sensitive to darkening the image, with lower scores at higher levels of distortion. The response to gaussian noise and salt-and-pepper noise is U-shaped, and the sensitivity to these types of noise is extremely high: low- and medium- distortion levels are given extremely poor skill ratings, whereas high distortion levels are given slightly less poor but still abysmal ratings. Images to which black rectangles have been added also show a different pattern: they are given the lowest rating at the smallest distortion level, when few small rectangles have been added, and are not penalized very strongly when much or all of the image has been covered up by rectangles}
  \label{fig-eoe}
\end{figure}

We undertake an analysis of skill rating's performance on distorted samples.
We skill rate SVHN images to which the following categories of distortion have been applied: Gaussian noise, salt-and-pepper noise, Gaussian blur, black rectangles, darkening, and lightening. (See \citep{Heusel2017} for an evaluation of Fr\'echet Inception Distance under similar distortions). The set of discriminators consists of experiments \dcgan{}, \wgan{}, and \feature{}.

One motivation for this analysis is to explore an observation which we make in Section~\ref{future}: namely, a GAN discriminator which is presented with a previously-unseen sample that looks nothing like the ``real'' samples it has seen, nor anything like the ``fake'' samples it has seen, does not have any particular incentive to label it as one or the other. We were interested to see whether images derived by distorting real samples might even be rated as ``realer than real''.

We make the following observations (see Figure~\ref{fig-eoe}):

\begin{enumerate}
  \item Skill rating is reasonably sensitive to lightening the image, and yet more sensitive to darkening the image, with lower scores at higher levels of distortion.
  \item Skill rating is not very sensitive to gaussian blur. We hypothesize this is because SVHN samples are often blurry, so failure to reject real samples that have been blurred is not an error.
  \item Gaussian noise and salt-and-pepper noise are rated poorly by the discriminators in this set at all distortion levels, although the curve is U-shaped, with the right-hand edge of the curve possibly seeming to flatten out. We hypothesize that discriminators have learned to be highly sensitive to this specific artifact because it emerges in the GAN training process: samples produced by generators early in training often appear to have high-frequency noise. We do not have a clear hypothesis at this time as to why medium levels of noise are given lower scores than high levels.
  \item When a few small rectangles have been added to the image, the samples are scored as somewhat worse than real samples, but when much or even \emph{all} of the image has been covered up by rectangles, the samples are penalized only slightly more than the lightened samples, which appear much less distorted at the same ``severity level''. We do not have a clear explanation for this, and speculate tentatively that this might be because sharp edges, like the transitions from the background to the black rectangles, are more typical of real samples than fake samples.
\end{enumerate}

\section{Appendix: Results with batchnorm discriminators}
\label{appendix-batchnorm}
As we mention in Section~\ref{compare}, we removed batchnorm from the discriminators and replaced it with pixelnorm, out of concern that our results would otherwise be hard to interpret. In an earlier version of these experiments, we kept batchnorm in the discriminators, and left them in ``training mode'' for the skill rating procedure - that is to say, rather than switching to use saved moving-average statistics, the discriminator continued to use the statistics of the incoming batch. We also did not add noise to the discriminator's input at training time in these experiments. We present those results here for the sake of interest.

\begin{figure}
  \centering
  \includegraphics[width=1.0\linewidth]{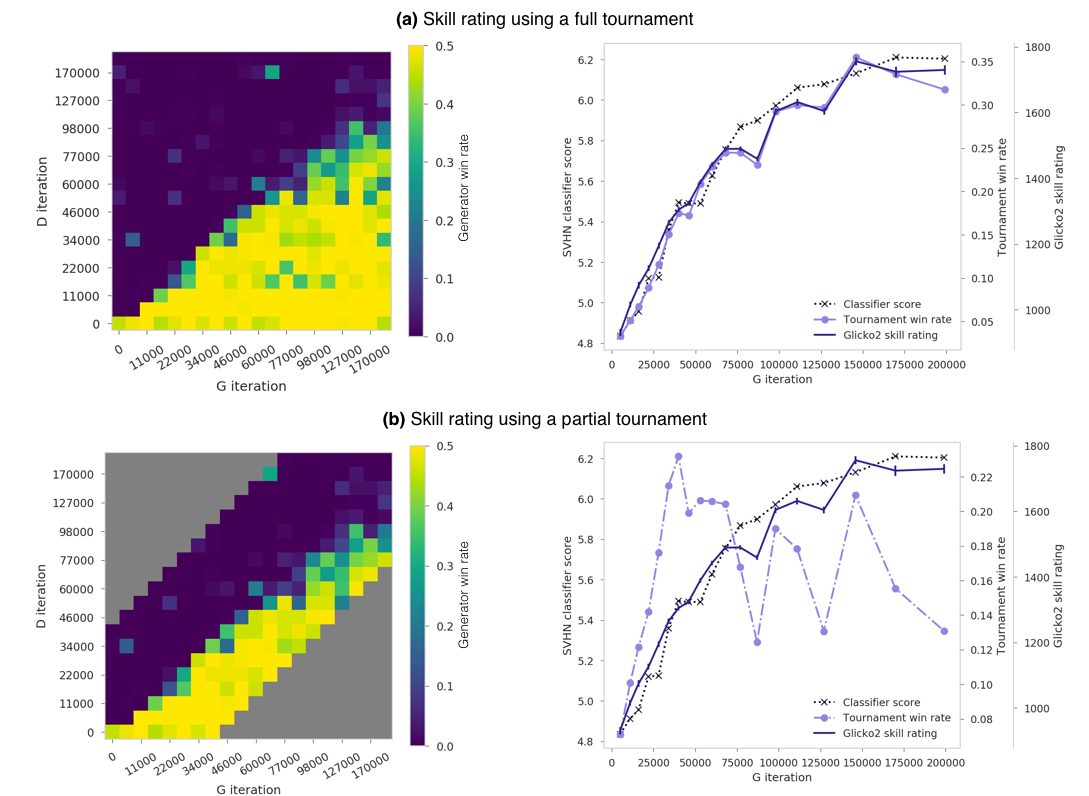}
    \caption{\textbf{Within-trajectory tournament outcomes for a DCGAN \emph{with} batchnorm.} This figure shows the same analysis as Figure~\ref{fig-monitor}, but for a DCGAN with batchnorm in the discriminator rather than pixelnorm.}
  \label{fig-appendix-batchnorm-monitor}
\end{figure}

\begin{figure}
  \centering
  \begin{subfigure}[b]{0.3\linewidth}
    \includegraphics[width=\linewidth]{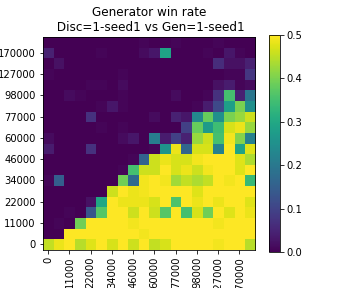}
    \caption{D=\dcgan{}-BN vs G=\dcgan{}-BN}
  \end{subfigure}
  \begin{subfigure}[b]{0.3\linewidth}
    \includegraphics[width=\linewidth]{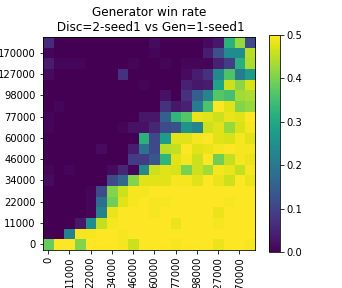}
    \caption{D=\feature{}-BN vs G=\dcgan{}-BN}
  \end{subfigure}
  \begin{subfigure}[b]{0.3\linewidth}
    \includegraphics[width=\linewidth]{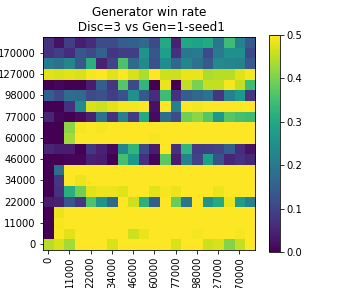}
    \caption{D=\wgan{}-BN vs G=\dcgan{}-BN}
  \end{subfigure}
  \\
  \begin{subfigure}[b]{0.3\linewidth}
    \includegraphics[width=\linewidth]{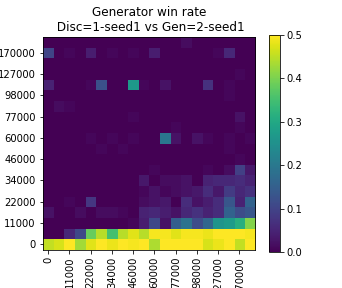}
    \caption{D=\dcgan{}-BN vs G=\feature{}-BN}
  \end{subfigure}
  \begin{subfigure}[b]{0.3\linewidth}
    \includegraphics[width=\linewidth]{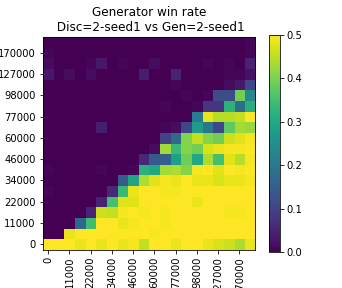}
    \caption{D=\feature{}-BN vs G=\feature{}-BN}
  \end{subfigure}
  \begin{subfigure}[b]{0.3\linewidth}
    \includegraphics[width=\linewidth]{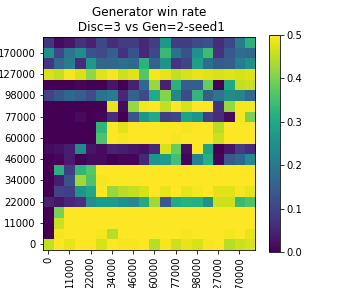}
    \caption{D=\wgan{}-BN vs G=\feature{}-BN}
  \end{subfigure}
  \\
  \begin{subfigure}[b]{0.3\linewidth}
    \includegraphics[width=\linewidth]{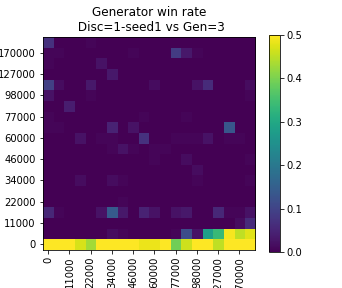}
    \caption{D=\dcgan{}-BN vs G=\wgan{}-BN}
  \end{subfigure}
  \begin{subfigure}[b]{0.3\linewidth}
    \includegraphics[width=\linewidth]{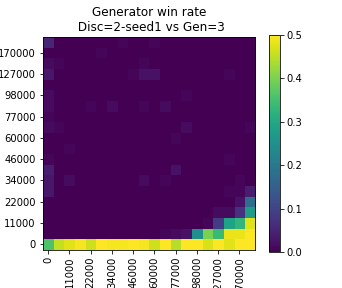}
    \caption{D=\feature{}-BN vs G=\wgan{}-BN}
  \end{subfigure}
  \begin{subfigure}[b]{0.3\linewidth}
    \includegraphics[width=\linewidth]{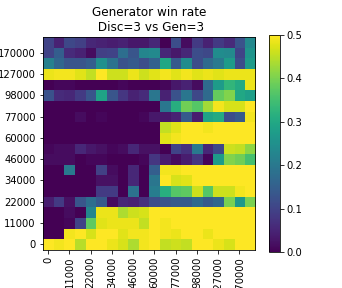}
    \caption{D=\wgan{}-BN vs G=\wgan{}-BN}
  \end{subfigure}
  \\
  \begin{subfigure}[b]{0.3\linewidth}
    \includegraphics[width=\linewidth]{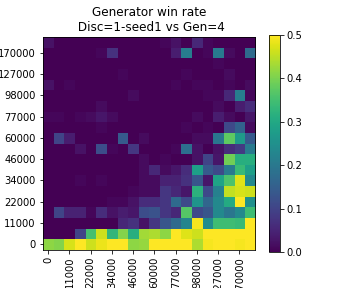}
    \caption{D=\dcgan{}-BN vs G=\conditional{}-BN}
  \end{subfigure}
  \begin{subfigure}[b]{0.3\linewidth}
    \includegraphics[width=\linewidth]{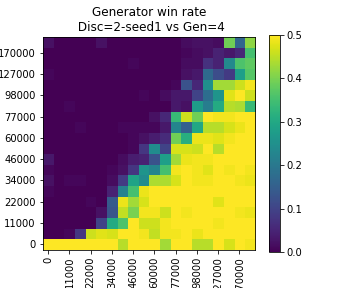}
    \caption{D=\feature{}-BN vs G=\conditional{}-BN}
  \end{subfigure}
  \begin{subfigure}[b]{0.3\linewidth}
    \includegraphics[width=\linewidth]{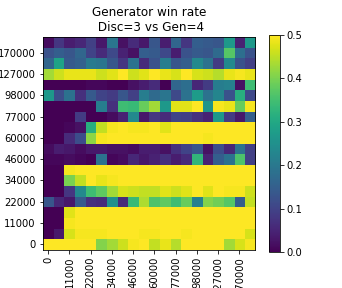}
    \caption{D=\wgan{}-BN vs G=\conditional{}-BN}
  \end{subfigure}
  \\
  \caption{\textbf{Tournament heatmaps \emph{with} batchnorm}. This figure shows the same analysis as in Figure~\ref{fig-appendix-winrate}, but with batchnorm in the discriminators. Note the substantially darker heatmaps than those in Figure~\ref{fig-appendix-winrate}, indicating a lower overall generator win rate with the batchnorm discriminators than with the pixelnorm discriminators in these experiments. We believe this darker pattern is indicative of experiments without noise added to the discriminator input, rather than a property that separates batchnorm from pixelnorm.}
  \label{fig-appendix-batchnorm-winrate}
\end{figure}

\begin{figure}
  \centering
  \begin{subfigure}[b]{0.24\linewidth}
    \includegraphics[width=\linewidth]{figs/samples_real_data_0.png}
    \caption{Real data samples\\\textbf{SR=1738}\\ CS=8.09}
  \end{subfigure}
  \begin{subfigure}[b]{0.24\linewidth}
    \includegraphics[width=\linewidth]{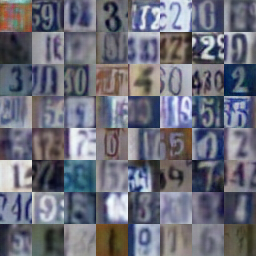}
    \caption{\dcgan{}-BN\\\textbf{SR=1610}\\ CS=6.27}
    \label{fig-samples-dcgan}
  \end{subfigure}
  \begin{subfigure}[b]{0.24\linewidth}
    \includegraphics[width=\linewidth]{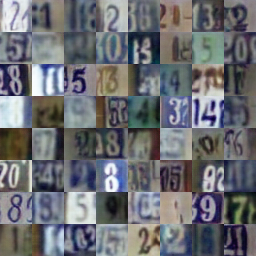}
    \caption{\conditional{}-BN\\\textbf{SR=1594}\\ CS=7.14}
    \label{fig-samples-conditional}
  \end{subfigure}
  \begin{subfigure}[b]{0.24\linewidth}
    \includegraphics[width=\linewidth]{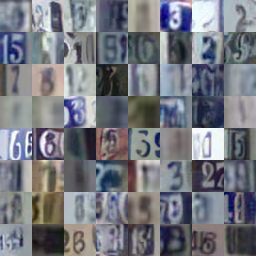}
    \caption{\pixelcnn{}\\\textbf{SR=1568}\\ CS=5.27}
    \label{fig-samples-pixelcnn}
  \end{subfigure}
  \\
  \begin{subfigure}[b]{0.24\linewidth}
    \includegraphics[width=\linewidth]{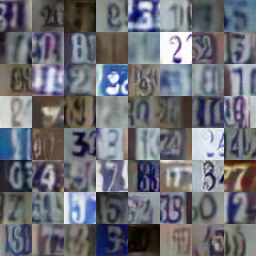}
    \caption{\feature{}-BN\\\textbf{SR=1511}\\ CS=6.08}
  \end{subfigure}
  \begin{subfigure}[b]{0.24\linewidth}
    \includegraphics[width=\linewidth]{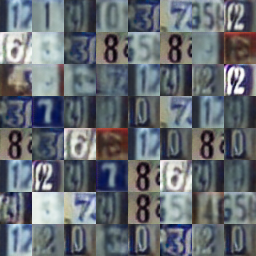}
    \caption{\eleven{}-BN\\\textbf{SR=1461}\\ CS=7.85}
  \end{subfigure}
  \begin{subfigure}[b]{0.24\linewidth}
    \includegraphics[width=\linewidth]{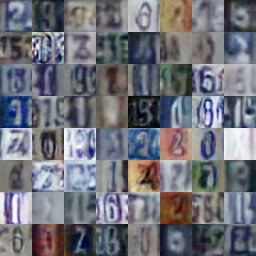}
    \caption{\wgan{}-BN\\\textbf{SR=1381}\\ CS=5.50}
  \end{subfigure}
  \begin{subfigure}[b]{0.24\linewidth}
    \includegraphics[width=\linewidth]{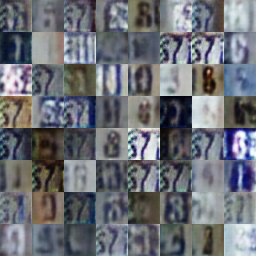}
    \caption{\dcgan{}-bad-lr-BN\\\textbf{SR=1361}\\ CS=4.67}
  \end{subfigure}

  \caption{\textbf{Samples from fully-trained generative models \emph{with} batchnorm}.}
  \label{fig-appendix-batchnorm-samples}
\end{figure}

Figure~\ref{fig-appendix-batchnorm-monitor} presents an alternate version of the within-trajectory heatmaps and scores in Figure~\ref{fig-monitor}.
Figure~\ref{fig-appendix-batchnorm-multi} presents an alternate version of the multi-experiment comparison scores in Figure~\ref{fig-multi}. Note that real data has been given a substantially higher rating than the best generated data in this experiment, whereas real data was rated much more closely to the best generated data in the pixelnorm experiments.
Figure~\ref{fig-appendix-batchnorm-winrate} presents full multi-experiment win rate heatmaps, as in Figure~\ref{fig-appendix-winrate}.
Figure~\ref{fig-appendix-batchnorm-samples} presents the samples from the batchnorm discriminator experiments.

\begin{figure}
  \centering
  \begin{subfigure}[b]{0.45\linewidth}
    \includegraphics[width=\linewidth]{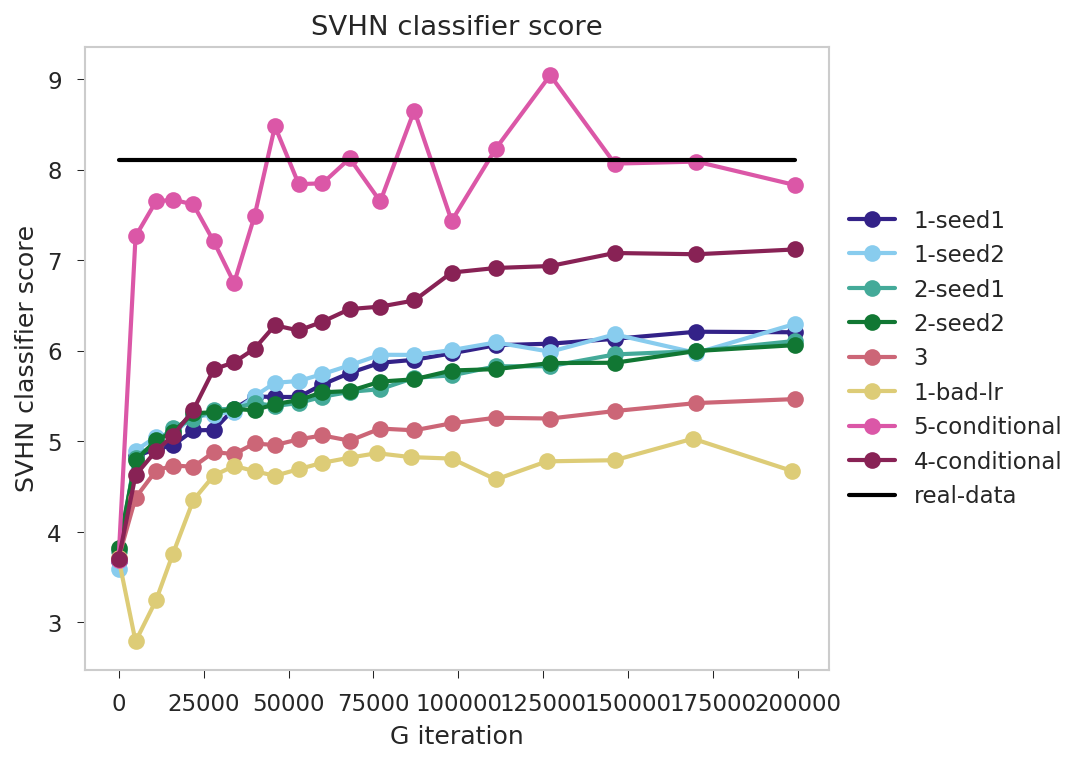}
  \end{subfigure}
  \begin{subfigure}[b]{0.45\linewidth}
    \includegraphics[width=\linewidth]{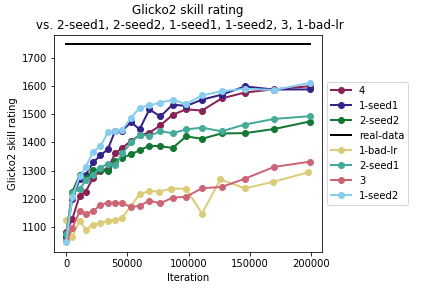}
  \end{subfigure}
\end{figure}

\begin{figure}
  \caption{\textbf{Multiple-trajectory tournament outcomes with batchnorm discriminators.} This figure shows the same analysis as Figure~\ref{fig-multi}, but with batchnorm in the discriminators.}
  \label{fig-appendix-batchnorm-multi}
\end{figure}

\captionsetup[subfigure]{width=1.0\textwidth}
\captionsetup[subfigure]{width=0.9\textwidth}

\captionsetup[subfigure]{width=1.0\textwidth}
\captionsetup[subfigure]{width=0.9\textwidth}

\section{Appendix: Architectures}

\subsection{SVHN task training procedure, model architectures, and hyperparameters}
\label{svhn_architecture}

Our SVHN GANs described in Section~\ref{compare} are based off the DCGAN architecture, loss function, and hyperparameters from \citet{Gulrajani2017}, with small modifications. To all GANs, we add noise to the generated and real samples at training time (standard deviation of 0.2), and we substitute pixelnorm instead of batchnorm in the discriminator only (with an epsilon of $1e-5$) \citep{Karras2017}. Experiment \dcgan{} is an otherwise-unmodified ordinary DCGAN. Experiment \wgan{} is a Wasserstein GAN \citep{Gulrajani2017}. Experiment \feature{} is a feature-matching GAN \citep{Salimans2016}. Experiment \conditional{} is a conditional GAN in which the label information is concatenated into the input to both generator and discriminator as an additional dimension. Experiment \eleven{} is a conditional GAN in which the discriminator makes an 11-way judgment: 10 real classes or ``fake''. All GAN models were trained for 200,000 steps with a learning rate of 0.0002 on both the generator and the discriminator. Experiment \pixelcnn{} is a PixelCNN++ \citep{Salimans2017}, trained using the code from \url{https://github.com/openai/pixel-cnn}, modified only to accept SVHN in place of CIFAR10.

We made no attempt to tune each model for its best possible performance, as it was advantageous for our purposes to allow sample quality to vary. As we emphasize in Section~\ref{compare}, our method is intended to compare the outcomes of individual experiments; we explicitly discourage an interpretation that we are comparing general algorithms.

\paragraph{Raw code for GAN architectures } All SVHN DCGAN variants used the architecture and hyperparameters described in the tensorflow code below. The different variants are defined by the flags in the code.

\begin{verbatim}
_HEIGHT, _WIDTH, _NUM_CHANNELS = [32, 32, 3]
WGAN_CRITIC_ITERS = 5


def _leaky_relu(x):
  return tf.maximum(0.2 * x, x)


def pixel_norm_nchw(x, eps=1e-8):
  return x * tf.rsqrt(tf.reduce_mean(tf.square(x), [1], keepdims=True) + eps)


def disc_inputs_with_labels(inputs, labels, scope, nplusone=False):
    height, width, _ = inputs.get_shape().as_list()[1:]
    # If fake_labels is not None, GAN is normal conditional.
    if nplusone:
      labels = None   # If nplusone, don't 'show' labels to D.
    if labels is not None:
      label_embedding = ops.linear.Linear(scope + '.labels', 10, height * width * 1, labels)
      label_embedding = _leaky_relu(label_embedding)
      label_embedding = tf.reshape(label_embedding, [-1, height, width, 1])
      inputs = tf.concat([inputs, label_embedding], axis=-1)
    return inputs


def ishaan_generator(z, fake_labels, is_training, stats_iter, scope, dim_z=128, add_to_collection=True):

    # If fake_labels is not None, GAN is conditional. Show fake_labels to G in that case.
    if fake_labels is not None:
      z = tf.concat([z, fake_labels], axis=-1)
      dim_z += fake_labels.get_shape().as_list()[-1]

    dim_g = 64
    output = ops.linear.Linear(scope + '.Input', dim_z, 4*4*4*dim_g, z)
    output = ops.batchnorm.Batchnorm(
        scope + '.BN1', [0], output, is_training=is_training, stats_iter=stats_iter)
    output = tf.nn.relu(output)
    output = tf.reshape(output, [-1, 4*dim_g, 4, 4])  # NCHW

    output = ops.deconv2d.Deconv2D(scope + '.2', 4*dim_g, 2*dim_g, 5, output)
    output = ops.batchnorm.Batchnorm(
        scope + '.BN2', [0,2,3], output, is_training=is_training, stats_iter=stats_iter)
    output = tf.nn.relu(output)

    output = ops.deconv2d.Deconv2D(scope + '.3', 2*dim_g, dim_g, 5, output)
    output = ops.batchnorm.Batchnorm(
        scope + '.BN3', [0,2,3], output, is_training=is_training, stats_iter=stats_iter)
    output = tf.nn.relu(output)

    output = ops.deconv2d.Deconv2D(scope + '.5', dim_g, 3, 5, output)

    output = tf.tanh(output)
    output = tf.reshape(output, [-1, _NUM_CHANNELS, _HEIGHT, _WIDTH])
    output = tf.transpose(output, [0, 2, 3, 1])  # move C back out to NHWC

    params = lib.params_with_name(scope, trainable_only=True)
    if add_to_collection:
      tf.add_to_collection(G_OUTPUT, output)
      tf.add_to_collection(G_PARAMS, params)
    return output, params


def pixelnorm_discriminator(inputs, labels, scope, add_to_collection=True, nplusone=False, eps=1e-8):

    inputs = disc_inputs_with_labels(inputs, labels, scope, nplusone)
    _, _, channels = inputs.get_shape().as_list()[1:]

    dim_d = 64
    output = tf.transpose(inputs, [0, 3, 1, 2])  # move C into NCHW
    output = ops.conv2d.Conv2D(scope + '.1', channels, dim_d, 5, output, stride=2)
    output = _leaky_relu(output)

    output = ops.conv2d.Conv2D(scope + '.2', dim_d, 2*dim_d, 5, output, stride=2)
    output = pixel_norm_nchw(output, eps)
    output = _leaky_relu(output)

    output = ops.conv2d.Conv2D(scope + '.3', 2*dim_d, 4*dim_d, 5, output,
                                      stride=2)
    output = pixel_norm_nchw(output, eps)
    output = _leaky_relu(output)

    output = tf.reshape(output, [-1, 4*4*4*dim_d])
    d_features = output  # for feature-matching

    if nplusone:
      output = ops.linear.Linear(scope + '.Output', 4*4*4*dim_d, 11, output)
    else:
      output = ops.linear.Linear(scope + '.Output', 4*4*4*dim_d, 1, output)
      output = tf.reshape(output, [-1])

    params = lib.params_with_name(scope, trainable_only=True)
    if add_to_collection:
      tf.add_to_collection(D_INPUT, inputs)
      tf.add_to_collection(D_OUTPUT, output)
      tf.add_to_collection(D_PARAMS, params)

    return output, params, d_features


def ns_discriminator_loss(d_on_data_logits, d_on_g_logits,
                          add_to_collection=True):
  loss = tf.reduce_mean(
    tf.nn.sigmoid_cross_entropy_with_logits(
      labels=tf.ones_like(d_on_data_logits), logits=d_on_data_logits) +
    tf.nn.sigmoid_cross_entropy_with_logits(
      labels=tf.zeros_like(d_on_g_logits), logits=d_on_g_logits))
  if add_to_collection:
    tf.add_to_collection(D_LOSS, loss)
  return loss


def ns_generator_loss(d_on_g_logits, add_to_collection=True):
  loss = tf.reduce_mean(tf.nn.sigmoid_cross_entropy_with_logits(
    labels=tf.ones_like(d_on_g_logits), logits=d_on_g_logits))
  if add_to_collection:
    tf.add_to_collection(G_LOSS, loss)
  return loss


def ns_train_op(cost, params, learning_rate, beta_1, collection=None):
  op = tf.train.AdamOptimizer(
    learning_rate, beta_1, 0.999).minimize(
    cost, var_list=params, colocate_gradients_with_ops=True)
  if collection:
    tf.add_to_collection(collection, op)
  return op


def nplusone_discriminator_loss(d_on_data_logits, d_on_g_logits,
                                real_labels, fake_labels,
                                add_to_collection=True):
  bsz = fake_labels.shape[0]

  augmented_real_labels = tf.zeros([bsz, 1])
  augmented_real_labels = tf.concat([real_labels, augmented_real_labels], -1)

  augmented_fake_labels = tf.ones([bsz, 1])
  augmented_fake_labels = tf.concat(
    [tf.zeros_like(fake_labels), augmented_fake_labels], -1)

  loss = tf.reduce_mean(
    tf.nn.softmax_cross_entropy_with_logits(
      labels=augmented_real_labels, logits=d_on_data_logits) +
    tf.nn.softmax_cross_entropy_with_logits(
      labels=augmented_fake_labels, logits=d_on_g_logits))
  if add_to_collection:
    tf.add_to_collection(D_LOSS, loss)
  return loss


def nplusone_generator_loss(d_on_g_logits, fake_labels, add_to_collection=True):
  bsz = fake_labels.shape[0]
  augmented_fake_labels = tf.zeros([bsz, 1])
  augmented_fake_labels = tf.concat([fake_labels, augmented_fake_labels], -1)

  loss = tf.reduce_mean(tf.nn.sigmoid_cross_entropy_with_logits(
    labels=augmented_fake_labels, logits=d_on_g_logits))
  if add_to_collection:
    tf.add_to_collection(G_LOSS, loss)
  return loss


def wgan_generator_loss(d_on_g_logits, add_to_collection=True):
  loss = -tf.reduce_mean(d_on_g_logits)
  if add_to_collection:
    tf.add_to_collection(G_LOSS, loss)
  return loss


def wgan_discriminator_loss(d_on_data_logits, d_on_g_logits,
                            add_to_collection=True):
  loss = tf.reduce_mean(d_on_g_logits) - tf.reduce_mean(d_on_data_logits)
  if add_to_collection:
    tf.add_to_collection(D_LOSS, loss)
  return loss


def wgan_train_op(cost, params, learning_rate, collection=None):
  op = tf.train.RMSPropOptimizer(learning_rate=learning_rate).minimize(
    cost, var_list=params)
  if collection:
    tf.add_to_collection(collection, op)
  return op


def wgan_clip_op(disc_params, add_to_collection=True):
  clip_ops = []
  for var in disc_params:
    clip_bounds = [-.01, .01]
    clip_ops.append(
      tf.assign(
        var,
        tf.clip_by_value(var, clip_bounds[0], clip_bounds[1])
        )
      )
  clip_disc_weights = tf.group(*clip_ops)
  if add_to_collection:
    tf.add_to_collection(CLIP_OP, clip_disc_weights)
  return clip_disc_weights


def feature_matching_generator_loss(d_g_features, d_data_features,
                                    add_to_collection=True):
  assert(len(d_g_features.shape) == 2)
  assert(len(d_data_features.shape) == 2)
  loss = tf.reduce_mean(tf.abs(
    tf.reduce_mean(d_g_features, axis=0) -
    tf.reduce_mean(d_data_features, axis=0)))
  if add_to_collection:
    tf.add_to_collection(G_LOSS, loss)
  return loss


def build_dcgan(dataset, batch_size=64, dim_z=128, learning_rate=2e-4,
                beta_1=0.5, graph=None, loss_variant='dcgan',
                eps=1e-8, ngrp=32, disc_noise=0.0,
                deterministic=False, tf_seed=TF_RNG_SEED,
                d_learning_rate=None, g_learning_rate=None):

  # Set default learning rates
  d_learning_rate = d_learning_rate or learning_rate
  g_learning_rate = g_learning_rate or learning_rate

  if graph is None:
    graph = tf.get_default_graph()
  if deterministic:
    tf.set_random_seed(tf_seed)

  with graph.as_default():
    # noise -> generated -> discriminator
    z = tf.random_normal([batch_size, dim_z])
    tf.add_to_collection(NOISE, z)

    real_data, real_labels = get_data_iterator(dataset, batch_size)
    if loss_variant in ['dcgan', 'feature', 'wgan']:
      real_labels = None
      fake_labels = None
    elif loss_variant in ['conditional', 'nplusone']:
      real_labels = tf.one_hot(real_labels, 10)
      fake_labels = tf.random_uniform(shape=[batch_size,], minval=0,
                                          maxval=10, dtype=tf.int32)
      fake_labels = tf.one_hot(fake_labels, 10)

    generator_output, g_params = ishaan_generator(
      z, fake_labels, is_training=None, stats_iter=None,
      scope=G_SCOPE, dim_z=dim_z)

    noisy_gen = generator_output + tf.random_normal(
      [batch_size, _HEIGHT, _WIDTH, _NUM_CHANNELS],
      mean=0.0, stddev=disc_noise)
    noisy_real = real_data + tf.random_normal(
      [batch_size, _HEIGHT, _WIDTH, _NUM_CHANNELS],
      mean=0.0, stddev=disc_noise)

    nplusone = (loss_variant == 'nplusone')

    d_on_g_logits, d_params, d_g_features = pixelnorm_discriminator(
      noisy_gen, fake_labels, scope=D_SCOPE, add_to_collection=True,
      nplusone=nplusone, eps=eps)
    d_on_data_logits, _, d_data_features = pixelnorm_discriminator(
      noisy_real, real_labels, scope=D_SCOPE, add_to_collection=False,
      nplusone=nplusone, eps=eps)

    # losses:
    if loss_variant == 'feature':
      g_loss = feature_matching_generator_loss(d_g_features, d_data_features,
                                               add_to_collection=True)
      d_loss = ns_discriminator_loss(d_on_data_logits, d_on_g_logits,
                                    add_to_collection=True)
    elif loss_variant == 'dcgan' or loss_variant == 'conditional':
      g_loss = ns_generator_loss(d_on_g_logits, add_to_collection=True)
      d_loss = ns_discriminator_loss(d_on_data_logits, d_on_g_logits,
                                    add_to_collection=True)

    elif loss_variant == 'wgan':
      g_loss = wgan_generator_loss(d_on_g_logits, add_to_collection=True)
      d_loss = wgan_discriminator_loss(d_on_data_logits, d_on_g_logits,
                                       add_to_collection=True)
    elif loss_variant == 'nplusone':
      g_loss = nplusone_generator_loss(d_on_g_logits, fake_labels, add_to_collection=True)
      d_loss = nplusone_discriminator_loss(d_on_data_logits, d_on_g_logits,
                                           real_labels, fake_labels,
                                           add_to_collection=True)

    # opts:
    if loss_variant in ['dcgan', 'feature', 'conditional', 'nplusone']:
      ns_train_op(cost=g_loss, params=g_params, learning_rate=g_learning_rate,
                    beta_1=beta_1, collection=G_OPT)

      ns_train_op(cost=d_loss, params=d_params, learning_rate=d_learning_rate,
                    beta_1=beta_1, collection=D_OPT)

    elif loss_variant == 'wgan':
      for (cost, params, collection) in [(g_loss, g_params, G_OPT),
                                         (d_loss, d_params, D_OPT)]:
        wgan_train_op(cost=cost, params=params, learning_rate=learning_rate,
                      collection=collection)

      wgan_clip_op(d_params)

\end{verbatim}

\subsection{"Gaussian Toy" task architecture and hyperparameters}
\label{toy_task_architecture}
For the ``Gaussian Toy'' task, we trained a small GAN (consisting of an MLP generator and an MLP discriminator with architectures described below) to estimate a 50-dimensional Gaussian. We additionally trained a separate Chekhov GAN. We show here the verbatim code from the Chekhov version. The vanilla toy GAN uses exactly the same architecture and training process, but without any past generators or discriminators.

\paragraph{Chekhov toy architecture}

\begin{verbatim}
class ChekhovToy(object):
  """A toy Chekhov GAN which estimates a dim-dimensional Gaussian"""

  def __init__(self, data_dir=DATA_DIR, batch_size=BATCH_SIZE, dim=DIM,
               np_seed=NP_RNG_SEED, tf_seed=TF_RNG_SEED, deterministic=False,
               queue_size=1, queue_spacing=1000, reservoir=False):
    self.data_dir = data_dir
    self.batch_size = batch_size
    self.dim = dim
    self.queue_size = queue_size
    self.queue_spacing = queue_spacing

    self.reservoir = reservoir

    self.graph = tf.Graph()
    self.sess = tf.Session(graph=self.graph)
    self._loaded_from = None

    with self.graph.as_default():
      if deterministic:
        self.rng = np.random.RandomState(np_seed)
        tf.set_random_seed(tf_seed)
      else:
        self.rng = np.random.RandomState(None)

      self._build_model()
      self.sess.run(tf.global_variables_initializer())


  def _build_generator(self, name):
    generator = MLP(name=name,
      layers=[Linear(self.dim, init_scale=.05)],
        input_shape=[self.batch_size, self.dim])
    return generator


  def _build_discriminator(self, name):
    discriminator = MLP(name=name,
      layers=[
          Linear(1200),
          ReLU(),
          Linear(1200),
          ReLU(),
          Linear(100),
          ReLU(),
          Linear(1)
      ],
      input_shape=[self.batch_size, self.dim])
    return discriminator


  def _build_generator_loss(self, d_on_g_logits):
    return tf.reduce_mean(tf.nn.sigmoid_cross_entropy_with_logits(
        labels=tf.ones_like(d_on_g_logits), logits=d_on_g_logits))


  def _build_discriminator_loss(self, d_on_data_logits, d_on_g_logits):
    return tf.reduce_mean(
        tf.nn.sigmoid_cross_entropy_with_logits(
            labels=tf.ones_like(d_on_data_logits), logits=d_on_data_logits) +
        tf.nn.sigmoid_cross_entropy_with_logits(
            labels=tf.zeros_like(d_on_g_logits), logits=d_on_g_logits))


  def _build_model(self):

    ##### Data ######

    # Generate samples from dim-dimensional Gaussians
    self._true_mu = tf.Variable(self.rng.randn(self.dim).astype("float32"), trainable=False)
    self._true_cov = tf.Variable(self.rng.randn(self.dim, self.dim).astype("float32"), trainable=False)

    self._true_cov = tf.matmul(self._true_cov, self._true_cov, transpose_a=True)
    true_cov_chol = tf.Variable(tf.transpose(tf.cholesky(self._true_cov)), trainable=False)
    true_z = tf.random_normal([self.batch_size, self.dim])
    self._true_samples = tf.matmul(true_z, true_cov_chol) + self._true_mu


    ##### Generator ######

    # Creating the live/trainable generator.
    generator = self._build_generator(name="G_live")

    W, b = generator.get_params()
    assert len(W.get_shape()) == 2
    assert len(b.get_shape()) == 1

    # Compute moments
    self._gan_cov = tf.matmul(W, W, transpose_a=True)
    self._gan_mu = tf.identity(b)

    # Get MMD error signal
    self._err = {}
    self._err['mmd'] = tf.maximum(
        tf.reduce_max(tf.abs(self._gan_cov - self._true_cov)),
        tf.reduce_max(tf.abs(self._gan_mu - self._true_mu)))
    self._err['max_cov_diff'] = tf.reduce_max(tf.abs(self._gan_cov - self._true_cov))
    self._err['max_mu_diff'] = tf.reduce_max(tf.abs(self._gan_mu - self._true_mu))
    self._err['mean_cov_diff'] = tf.reduce_mean(tf.abs(self._gan_cov - self._true_cov))
    self._err['mean_mu_diff'] = tf.reduce_mean(tf.abs(self._gan_mu - self._true_mu))
    self._err['mean_sq_cov_diff'] = tf.reduce_mean(tf.square(self._gan_cov - self._true_cov))
    self._err['mean_sq_mu_diff'] = tf.reduce_mean(tf.square(self._gan_mu - self._true_mu))

    z = tf.random_normal([self.batch_size, self.dim])
    self._gan_samples = generator.fprop(z)


    ##### Discriminator #####

    # Creating the live/trainable discriminator
    discriminator = self._build_discriminator(name="D_live")
    d_on_data_logits = tf.squeeze(discriminator.fprop(self._true_samples))
    d_on_g_logits = tf.squeeze(discriminator.fprop(self._gan_samples))

    # Discriminate placeholder
    self._discriminate_input = tf.placeholder(tf.float32, shape=(self.batch_size, self.dim))
    self._discriminate_output = discriminator.fprop(self._discriminate_input)


    ##### Past G/D queue #####

    # Discriminator queue: old D's classify the current G's samples
    self._curr_g_old_d_losses = []
    self._d_queue_assign_ops = []

    for i in range(self.queue_size):
      past_d_i = self._build_discriminator(name="D_queue_{}".format(i))

      past_d_on_curr_g_logits = tf.squeeze(past_d_i.fprop(self._gan_samples))
      self._curr_g_old_d_losses.append(self._build_generator_loss(
        d_on_g_logits=past_d_on_curr_g_logits))

      self._d_queue_assign_ops.append(past_d_i.assign_params(
        discriminator.get_params(collapse=False)))

    # Generator queue: the current D classifies old G's samples
    self._curr_d_old_g_losses = []
    self._g_queue_assign_ops = []

    for i in range(self.queue_size):
      past_g_i = self._build_generator(name="G_queue_{}".format(i))
      z_i = tf.random_normal([self.batch_size, self.dim])
      past_samples_i = past_g_i.fprop(z_i)
      d_on_past_g_logits = tf.squeeze(discriminator.fprop(past_samples_i))
      self._curr_d_old_g_losses.append(self._build_discriminator_loss(
          d_on_g_logits=d_on_past_g_logits,
          d_on_data_logits=d_on_data_logits))

      self._g_queue_assign_ops.append(past_g_i.assign_params(
        generator.get_params(collapse=False)))


    ##### Loss & optimizers ######

    # Compute d_loss and g_loss
    self._d_curr_loss = self._build_discriminator_loss(
        d_on_data_logits=d_on_data_logits,
        d_on_g_logits=d_on_g_logits)
    self._g_curr_loss = self._build_generator_loss(d_on_g_logits)

    # Build a whole list of optimizers, depending on how full the queue is...
    self._d_losses = []
    self._d_opts = []
    self._g_losses = []
    self._g_opts = []
    for i in range(self.queue_size + 1):
      _d_loss_i = tf.reduce_mean(self._curr_d_old_g_losses[:i] + [self._d_curr_loss])
      _g_loss_i = tf.reduce_mean(self._curr_g_old_d_losses[:i] + [self._g_curr_loss])

      self._d_losses.append(_d_loss_i)
      self._g_losses.append(_g_loss_i)

      # NB: this does NOT include the L2 regularizer from Chekhov GAN
      d_optimizer = tf.train.AdamOptimizer(learning_rate=1e-3)
      g_optimizer = tf.train.AdamOptimizer(learning_rate=1e-3)

      _d_opt_i = d_optimizer.minimize(_d_loss_i, var_list=discriminator.get_params())
      _g_opt_i = g_optimizer.minimize(_g_loss_i, var_list=generator.get_params())

      self._d_opts.append(_d_opt_i)
      self._g_opts.append(_g_opt_i)


    ##### Saver ######
    self._saver = tf.train.Saver(max_to_keep=None)


  def load(self, modelID):
    if self._loaded_from != modelID:
      with self.sess.as_default():
        with self.graph.as_default():
          self._saver.restore(
              self.sess, os.path.join(logdir(modelID.experiment),
                                      modelID.experiment + "-{}".format(modelID.iteration)))
          self._loaded_from = modelID

  def save(self, experiment_name, global_step):
    assert(experiment_name)
    self._saver.save(
      self.sess,
      os.path.join(logdir(experiment_name), experiment_name),
      global_step=global_step,
      write_meta_graph=(global_step == 0))


  def _should_evict_now(self, i):
    if self.queue_size == 0 or i == 0:
      return False

    else:
      if self.reservoir:
        return (i < self.queue_size or
                np.random.random() > (self.queue_size / i))
      else:
        return i % self.queue_spacing


  def train(self, steps=16000, experiment_name=None, save=False):
    with self.sess.as_default():
      with self.graph.as_default():
        row_format_s = "{:^10} | {:^16} | {:^16} | {:^16} | {:^16} | {:^16}"
        row_format = "{:^10} | {:^16.1e} | {:^16.1e}| {:^16.1e}| {:^16.1e} | {:^16.4f}"
        print(row_format_s.format("Iteration", "D total loss", "D curr loss",
                                  "G total loss", "G curr loss", "MMD"))
        print("=" * 98)

        queue_idx = 0
        queue_full = False
        for i in xrange(steps + 1):

          # Update queues if it's time to do so
          if self._should_evict_now(i):
            self.sess.run(self._d_queue_assign_ops[queue_idx])
            self.sess.run(self._g_queue_assign_ops[queue_idx])

            if not queue_full and (queue_idx + 1) == self.queue_size:
              queue_full = True

            if queue_full and self.reservoir:
              queue_idx = np.random.randint(self.queue_size)  # if reservoir sampling into a full queue, evict randomly
            else:
              queue_idx = (queue_idx + 1) % self.queue_size  # otherwise, march through in an orderly fashion

          # Choose the loss function - if the queue isn't full, don't backprop through the unused slots
          if not queue_full:
            loss_idx = queue_idx
          else:
            loss_idx = self.queue_size

          # Print out
          if i % 1000 == 0:
            # Eval
            d_loss, d_curr_loss, g_loss, g_curr_loss, err = self.sess.run([
                self._d_losses[loss_idx],
                self._d_curr_loss,
                self._g_losses[loss_idx],
                self._g_curr_loss,
                self._err
            ])
            print(row_format.format(i, d_loss, d_curr_loss, g_loss, g_curr_loss, err))

            # Save
            if save:
              self.save(experiment_name, i)


          # Train
          self.sess.run([self._d_opts[loss_idx], self._g_opts[loss_idx]])

\end{verbatim}

\end{document}